\documentclass{article}

\usepackage{microtype}
\usepackage{graphicx}
\usepackage{subfigure}
\usepackage{booktabs} % for professional tables

\usepackage{hyperref}

\usepackage[accepted]{icml2025}

\usepackage{amsmath}
\usepackage{amssymb}
\usepackage{mathtools}
\usepackage{amsthm}

\usepackage[capitalize,noabbrev]{cleveref}

\theoremstyle{plain}

\theoremstyle{definition}

\theoremstyle{remark}

\usepackage[textsize=tiny]{todonotes}

\newif\ifdraft
\drafttrue

\ifdraft
\newcommand{\gal}[1]{{\color{orange}{[Gal: #1]}}}

\newcommand{\PF}[1]{{\color{red}{\bf PF: #1}}}

\newcommand{\AS}[1]{{\color{blue}{\bf AS: #1}}}

\newcommand{\ND}[1]{{\color{cyan}{\bf ND: #1}}}
\newcommand{\nd}[1]{{\color{cyan} #1}}
\newcommand{\DOn}[1]{{\color{green}{\bf DOn: #1}}}

\else
\newcommand{\PF}[1]{}

\newcommand{\AS}[1]{}

\newcommand{\ND}[1]{}
\newcommand{\nd}[1]{#1}
\newcommand{\DOn}[1]{}

\newcommand{\gal}[1]{}

\fi

\newcommand{\parag}[1]{\vspace{-1mm}\paragraph{#1}}

\newcommand{\bzer}{\mathbf{0}}

\newcommand{\bx}{\mathbf{x}}
\newcommand{\by}{\mathbf{y}}

\definecolor{gray}{rgb}{0.6, 0.6, 0.6} % Adjust the RGB values as needed% You can adjust 0.3 to be lighter or darker

\newcommand{\itt}[0]{IT$^3$}

\definecolor{bleudefrance}{RGB}{49,140,231}

\icmltitlerunning{IT³: Idempotent Test-Time Training}

\begin{document}

\twocolumn[
\icmltitle{\itt{}: Idempotent Test-Time Training}



\icmlsetsymbol{equal}{*}

\begin{icmlauthorlist}
\icmlauthor{Nikita Durasov}{equal,epfl}
\icmlauthor{Assaf Shocher}{equal,nv}
\icmlauthor{Doruk Oner}{bu}
\icmlauthor{Gal Chechik}{nv}
\icmlauthor{Alexei A. Efros}{ubc}
\icmlauthor{Pascal Fua}{epfl}
\end{icmlauthorlist}

\icmlaffiliation{epfl}{CVLAB, EPFL}
\icmlaffiliation{nv}{NVIDIA}
\icmlaffiliation{bu}{NeuraVision Lab, Bilkent University}
\icmlaffiliation{ubc}{UC Berkeley}

\icmlcorrespondingauthor{Nikita Durasov}{nikita.durasov@nvidia.com}

\icmlkeywords{Machine Learning, ICML}

\vskip 0.3in
]



\printAffiliationsAndNotice{This work was supported in part by the Swiss National Science Foundation. \icmlEqualContribution}  % leave blank if no need to mention equal contribution

% !TEX root = ../top.tex
% !TEX spellcheck = en-US

\begin{abstract}

Deep learning models often struggle when deployed in real-world settings due to distribution shifts between training and test data. While existing approaches like domain adaptation and test-time training (TTT) offer partial solutions, they typically require additional data or domain-specific auxiliary tasks. We present Idempotent Test-Time Training (IT$^3$), a novel approach that enables on-the-fly adaptation to distribution shifts using only the current test instance, without any auxiliary task design. Our key insight is that enforcing idempotence---where repeated applications of a function yield the same result---can effectively replace domain-specific auxiliary tasks used in previous TTT methods. We theoretically connect idempotence to prediction confidence and demonstrate that minimizing the distance between successive applications of our model during inference leads to improved out-of-distribution performance. Extensive experiments across diverse domains (including image classification, aerodynamics prediction, and aerial segmentation) and architectures (MLPs, CNNs, GNNs) show that IT$^3$ consistently outperforms existing approaches while being simpler and more widely applicable. Our results suggest that idempotence provides a universal principle for test-time adaptation that generalizes across domains and architectures.
\vspace{-2mm}
\begin{center}
\ \  \ \ \href{https://icml.cc/virtual/2025/poster/45551}{\textcolor{bleudefrance}{\texttt{poster}}}\ \ /
\ \ \href{https://github.com/nikitadurasov/ittt}{\textcolor{bleudefrance}{\texttt{code}}}\ \ /
\ \ \href{https://nikitadurasov.github.io/projects/ittt/video.html}{\textcolor{bleudefrance}{\texttt{video}}}\ \ /
\ \ \href{https://nikitadurasov.github.io/projects/ittt/}{\textcolor{bleudefrance}{\texttt{web}}}
\end{center}
\end{abstract}
\vspace{-4mm}

\begin{figure*}[htbp]
  \centering
  \includegraphics[width=0.9\linewidth]{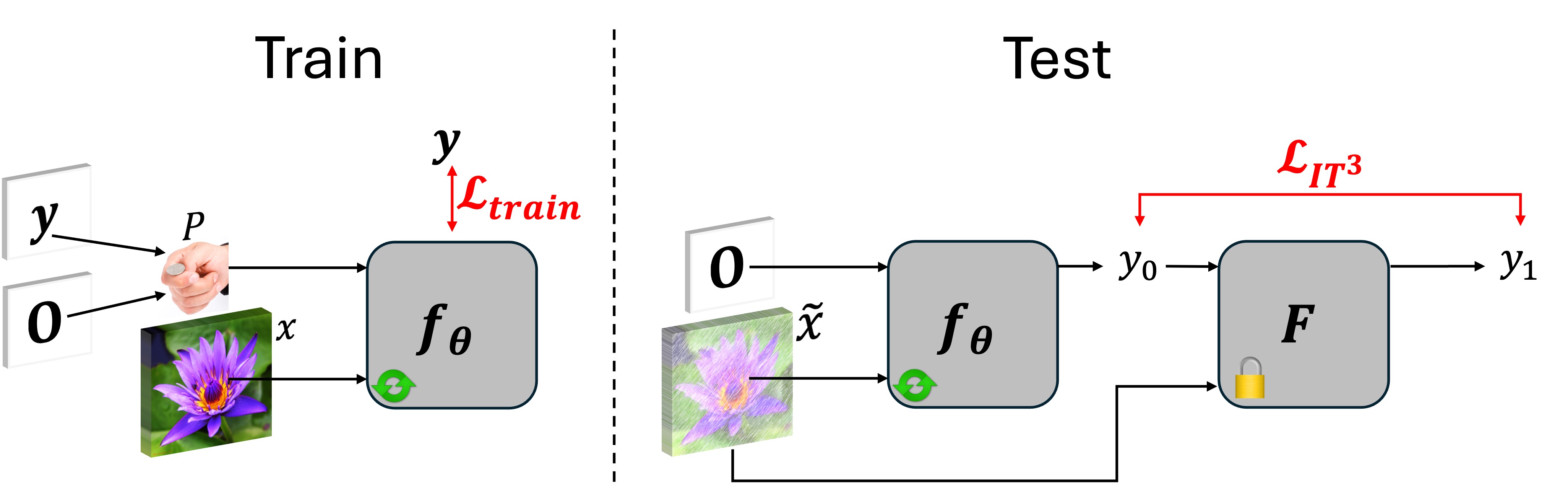}
  \vspace{-1mm}
  \caption{\small {\bf Idempotent Test-Time Training (\itt{}) approach.}  During training (left), the model $f_{\theta}$ is trained to predict the label $y$ with or without $y$ given to it as input. At test time (right), when given a corrupted input, the model is sequentially applied. It then briefly trains with the objective of making $f_{\theta}(\bx,\cdot)$ to be idempotent using only the current test input.}
  \label{fig:method}
\end{figure*}

\section{Introduction}

Supervised learning methods, while powerful, typically assume that training and test data come from the same distribution. Unfortunately, this is rarely true in practice. Data encountered by systems operating in the real world often differs substantially from what they were trained on due to data distribution shifts over time or other changes in the environment. This inevitably degrades performance, even in state-of-the-art models~\citep{Recht18, Hendrycks21b, Yao22a}. Machine learning systems used in production not only need to adapt to distribution shifts but also must do so on-the-fly using very limited data.

Thus, in this work, we focus on adapting to distribution shifts on-the-fly using only the current test instance or batch, without access to any additional labeled or unlabeled data during inference. During training, the model only has access to the base distribution training data, without any knowledge of the test distribution, which may be different. 

Adversarial robustness and domain adaptation address related challenges. However, they typically require additional data either during training or inference, and sometimes rely on specific assumptions about the nature of the shift. While effective when such additional knowledge is available, they are not designed for immediate, instance-level adaptation and are not applicable in the scenario we envision. Test-Time Training (TTT)~\citep{Sun20c} offers a promising alternative by adapting the model during inference by performing an auxiliary self-supervised task on each test sample. This enables the model to handle corrupted and Out-of-Distribution (OOD) data using only the current test instance or batch, without access to anything else. However, TTT requires performing an auxiliary task specific to the data modality, such as orientation prediction or inpainting for image data~\citep{Gandelsman22}. And defining an appropriate auxiliary task is not straightforward in general. 

In this paper, we argue that enforcing {\it idempotence} can profitably replace the auxiliary tasks in TTT and yields an approach we dub \itt{} that  is a versatile and powerful while generalizing well across  domains and architectures.  

% More specifically, let $f$ be a deep network that takes as input a vector $\bx$ and a second auxiliary variable that can either be the ground truth label $\by$  corresponding to $\bx$ or a neutral uninformative signal $\mathbf{0}$. In~\cite{Durasov24b}, it was shown that if such a network is trained so that $f(\bx,\mathbf{0})=f(\bx,\by)=\by$, then at test time the distance  $||f(\bx, f(\bx, \bzer)) - f(\bx, \bzer)||$  correlates strongly with the prediction error. Here, we propose to go one step further and to train $f$ at test time to minimize this distance. In effect, we want to make  \(f(\bx, \cdot)\) \textit{idempotent} for all $\bx$, meaning that it can be applied sequentially without changing the result beyond the initial  application.  This can be understood as a generalization of orthogonal projection in linear spaces to non-linear settings.

More specifically, let $f$ be a deep network that takes as input a vector $\bx$ and a second auxiliary variable that can either be the ground truth label $\by$  corresponding to $\bx$ or a neutral uninformative signal $\mathbf{0}$. In~\cite{Durasov24b}, it was shown that if such a network is trained so that $f(\bx,\mathbf{0})=f(\bx,\by)=\by$, then at test time the distance  $||f(\bx, f(\bx, \bzer)) - f(\bx, \bzer)||$  correlates strongly with the prediction error.
What if, at test time, we could actively minimize this distance whenever we encounter an OOD instance? Could we ``pull it'' into the distribution? \itt{} uses this distance as a loss for TTT sessions. Unfolding $y_0,y_1$ we obtain:  $||f(\bx, f(\bx, \mathbf{0})) - f(\bx, \mathbf{0})||$.  This makes \(f(\bx, \cdot)\) \textit{idempotent}. Fortunately, it has been shown that, while not trivial, training a model to achieve idempotence is feasible~\citep{Shocher24}. See further discussion in Appendix \ref{app:ign}.

This yields a generic  method that does not rely on any specific domain properties. This is in contrast to prior TTT methods that rely on a domain specific auxiliary task. By leveraging the universal property of idempotence, \itt{} can adapt OOD test inputs on-the-fly across various domains, tasks and architectures.
This includes image classification with corruptions, aerodynamic predictions for airfoils and cars, tabular data with missing information, age prediction from faces, and large-scale aerial photo segmentation, using MLPs, CNNs, or GNNs.

\section{Related Work}
\label{sec:related}

In essence, \itt{} relies on \textit{idempotence} to generalize \textit{Test-Time Training}. We briefly review these two fields. 

\subsection{Test-Time Training (TTT)}

The idea of leveraging test data for model adaptation dates back to methods like transductive learning ~\citep{Gammerman98}. Early approaches, such as transductive SVMs~\citep{Collobert06} and local learning~\citep{Bottou92}, aimed to adapt predictions for specific test samples rather than generalizing across unseen data.

Training neural networks solely on single test instances, without pre-training, has been demonstrated in the "deep internal learning" line of work, for many image enhancement tasks~\citep{Shocher18, Gandelsma19} and single image generative models~\citep{Shocher19, Shaham19}.

\textbf{Distribution Shifts}: TTT has emerged as a solution to the problem of generalization under distribution shifts. Using a pre-trained network and at test-time refining on a single instance each time. In the foundational work of~\citet{Sun20c}, the model is adjusted in real-time  by solving an auxiliary self-supervised task, such as predicting image rotations, on each test sample. This on-the-fly adaptation has proven good at improving robustness on corrupted and Out-Of-Distribution (OOD) data. As the self-supervised learning methods became more efficient~\citep{He22a}, they could be exploited for TTT~\citep{Gandelsman22}.
Extensions such as TTT++~\citep{Liu21f} require access to the entire test set. TENT~\citep{Wang21b} adapts during inference at the batch level, based on batch entropy, but cannot be applied to single instances or very small batches. Moreover, it relies on updating the model's normalization layers, making it architecture dependent. Another problem that many existing TTT approaches face is their task dependency, meaning they are designed to work for a specific data types, which is almost always image classification. The recent method ActMAD~\cite{Mirza23} addresses this limitation by pulling the mean and variance of data embeddings closer to those of the training data, improving predictions on out-of-distribution or corrupted data. While its effectiveness has been mostly demonstrated on image data, the approach has the potential to be applied to other types of data as well and we will use as one of the baselines we compare against. In~\citep{park2024test}, the authors propose a test-time adaptation method for depth completion, fine-tuning a specific adaptor layer using a consistency loss between two predictions. However, the approach is tailored to depth completion and not applicable to other tasks.

\textbf{TTT vs. Test Time Adaptation (TTA)}:
TTT operates per-instance, with no assumption that future test data will be similar. In contrast, TTA~\cite{Liang24a}  adapts using a \emph{large} test set from the same shift (Batch/Domain TTA), or assuming correlation between instances (Online TTA). Most previous work~\cite{Sun20c,Gandelsman22} have thus treated TTA and TTT as distinct paradigms rather than direct competitors. TTA exploits abundant test data but cannot tune the training process, while TTT shapes training but handles every test instance in isolation. Each approach suits different scenarios.

\subsection{Idempotence in Deep Learning}

Idempotence, a concept rooted in mathematics and functional programming, refers to an operation whose repeated application yields the same result as a single application. Mathematically, for a function $f$, being idempotent means
\begin{equation}
f(f(x)) = f(x), \quad \forall x \; .
\end{equation}
In other words, applying the function multiple times has no effect beyond the first one. 
In the context of linear operators, idempotence corresponds to orthogonal projection. Over \( \mathbb{R}^n \), these are matrices \( A \) that satisfy \( A^2 = A \), with eigenvalues of either 0 or 1; they can be interpreted as geometrically preserving certain components while nullifying others.
This concept was recently applied in generative modeling. Idempotent Generative Network (IGN)~\citep{Shocher24} is a generative model that maps data instances to themselves, \(f(x)=x\), and maps latents to targets that also map to themselves, \(f(f(z))=f(z)\). It was shown to 'project' corrupted images onto the data manifold, effectively removing the corruptions without prior knowledge of the degradation.

% For example,  Idempotent Generative Networks (IGN)~\citep{Shocher24} are a recent advance in this area. It leverages the concept of idempotence to train generative models. By enforcing idempotent mappings during the generative process, IGN ensures that the generative model can stably reconstruct its inputs, even under noisy or corrupted conditions. This property helps mitigate issues with distributional shifts, a core challenge in deploying models under real-world conditions where test distributions often differ from training data.
%
Energy-Based Models (EBMs; \cite{Ackley85}) offer a related perspective by defining a function \( f \) that assigns energy scores to inputs, with higher energy indicating less desirable or likely examples, and lower energy indicating those that fit the model well. IGN introduces a similar concept but frames it differently: Instead of \( f \) directly serving as the energy function, the energy is implicitly defined via the difference \( \delta(y) = D(f(y), y) \), where \( D \) measures the distance between the model’s prediction and its input. In this framework, training \( f \) to be idempotent minimizes \( \delta(f(z)) \), pushing the model toward a low-energy configuration where its outputs remain stable under repeated applications. Thus, \( f \) can be interpreted as a transition operator that drives high-energy inputs toward a low-energy, stable domain, reducing the need for separate optimization procedures to find the energy minimum.

In concurrent work, the ZigZag method has first been proposed and then extended to recursive networks~\citep{Durasov24a,Durasov24b}. It introduces idempotence as a means to assess uncertainty in neural network predictions. ZigZag operates by recursively feeding the model’s predictions back as inputs, allowing the model to refine its outputs. The consistency between successive predictions acts as an uncertainty metric, where stable, unchanged outputs indicate higher confidence, while divergent predictions signal uncertainty or out-of-distribution (OOD) data. Unlike popular sampling-based uncertainty estimation methods~\citep{Gal16a, Lakshminarayanan17, Wen2020, Durasov21a}, ZigZag does not require many forward passes or complex sampling, making it more computationally efficient for real-time applications.

%--------
% OLD
%--------

% However, the effectiveness of TTT often hinges on the alignment between the auxiliary task and the test distribution. If the self-supervised task is poorly chosen, the model may not adapt appropriately. For instance, simple tasks like image rotations may fail to capture the complexity of distribution shifts in more intricate domains~\citep{Gandelsman22}. Additionally, TTT tends to be domain-specific, limiting its application to tasks where the auxiliary task remains relevant. While TTT has been successfully applied in domains such as image classification and medical imaging, generalizing it across various tasks, architectures, and data types remains an open challenge.

% Recent extensions of TTT have aimed at broadening its applicability. TTT++~\citep{Liu21f} and TENT~\citep{Wang20k} relax the single-sample assumption by adapting to entire batches of test samples at once. These methods yield improvements when access to multiple test samples is available. However, such extensions still require careful tuning of the auxiliary task and the availability of multiple samples, which may not always be practical. Additionally, these methods introduce a computational overhead by performing optimization steps during inference, posing challenges for real-time applications and resource-constrained environments. \PF{Aren't we doing the same and suffering from the same overhead?}

% \input{tex/idem_to_ood}
% !TEX root = ../top.tex
% !TEX spellcheck = en-US

\section{Method}

Given a pre-trained model,  \itt{} aims to dynamically adapt its weights at inference time using Test-Time Training (TTT) to reduce uncertainty and handle Out-of-Distribution (OOD) instances. As discussed above, other TTT approaches rely on satisfying domain specific auxiliary tasks to achieve this. Instead, we rely on the model we train being idempotent on the training set and adapt its weights at inference time to approach idempotence on the test set as new samples are being fed to it. This pulls  the representations of OOD inputs back into the distribution of the training data and improves the model's performance on corrupted or OOD instances.

In this section, we describe how we make our models idempotent for training set samples, how we use the idempotence loss for TTT during inference, and how we adapt the algorithm for online scenarios.   

\subsection{Making the Network Idempotent at Training Time}
\label{ssec:zigzag-train}

\begin{figure}[ht!]
    \centering
    \includegraphics[width=0.49\textwidth]{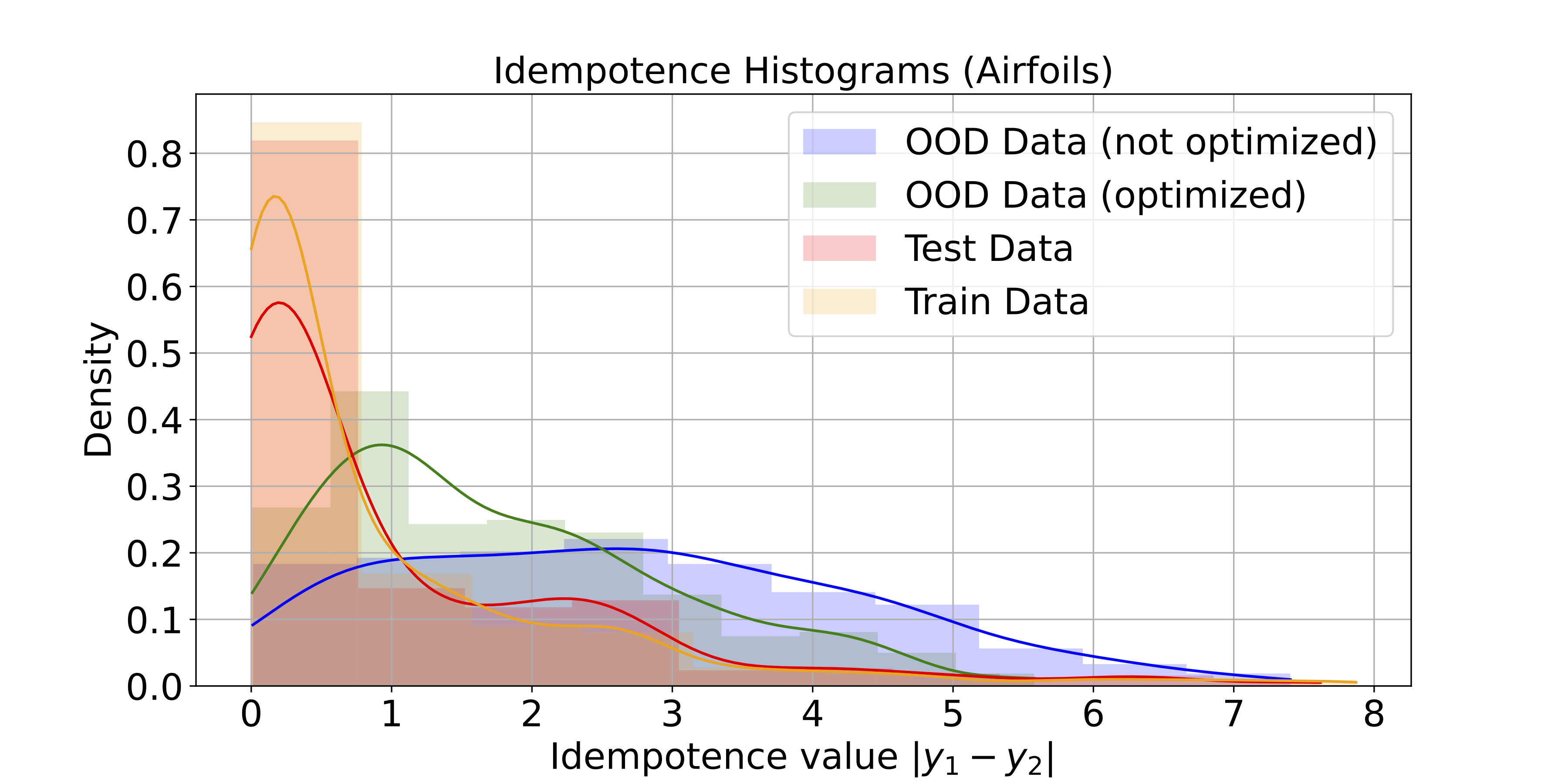}
    \vspace{-8mm}
    \caption{\small \textbf{Idempotence vs. Out-of-Distributionness}: We plot the distribution of idempotence errors, measured by the distance $| y_1 - y_2 |$ in Eq.\ref{eq:zigzag}, for training, test, and OOD data. For OOD samples, we show the errors both before and after minimizing them globally. OOD samples exhibit significantly larger idempotence errors, which decrease after optimization. Figuratively, \itt{} pushes the OOD representations to be more similar to those of the training distribution. In Sec.\ref{sec:experiments}, we show that this reduction yields improved performance.}
    \label{fig:idempotence_effect}
\end{figure}

Let $f_{\theta}$ be a generic network with weights $\theta$ that takes an input $\bx$. We wish to deploy in an environment where the statistical distribution of the samples it receives may change over time. To this end, as in ZigZag~\cite{Durasov24a}, we modify slightly its input layer so that it can accept a second argument that be either $\by$, the desired output of  the network given input $\bx$, or a neutral uninformative signal $\bzer$. During the initial training, we  minimize the supervised loss
\begin{align}
\mathcal{L}_{\text{train}} = \| f_{\theta}(\bx, \by) - \by \| + \| f_{\theta}(\bx, \bzer) - \by \| \; ,
\label{eq:train-loss}
\end{align}
as depicted on the left side of Fig.~\ref{fig:method}. This enforces
\begin{align}
\by_0 &= f_{\theta}(\bx,\bzer)  \approx \by \; ,   \\
\by_1 &= f_{\theta}(\bx,\by_0)  \approx  f_{\theta}(\bx,\by) \approx \by \; , \label{eq:zigzag} \\
&\Rightarrow f_{\theta}(\bx, f_{\theta}(\bx,\bzer)) \approx f_{\theta}(\bx,\bzer) \nonumber \; .
\end{align}
In other words, $\theta$ has been adjusted so that function $f_{\theta}(\bx,\cdot)$ is as idempotent as possible for all $\bx$ in the training set. Of course, when $\bx$ is comes from the test set, there is no guarantee of that because there may be distribution shift between the two sets. In~\cite{Durasov24a}, it is shown that the deviation from the equality of Eq.~\ref{eq:zigzag} expressed as $\| f_{\theta}(\bx, f_{\theta}(\bx,\bzer)) - f_{\theta}(\bx,\bzer) \|$ correlates strongly with the accuracy of the prediction and can be used to detect testing samples that are out-of-distribution with respect to the training set. Fig.~\ref{fig:idempotence_effect} illustrates this in the case of a network trained to predict the lift-over-drag ratio (L/D) of a 2D airfoil. In other words, here, $\bx$ is a 2D outline representing an airfoil and the output $y$ is expected the corresponding L/D.

% !TEX root = ../top.tex
% !TEX spellcheck = en-US

%therefore, given an in-distribution  \( (\bx, \by) \) pair, the model learns to return \( \by_0 = \by_1 \). In other words, for any \( \bx \) in the training set, \( f_{\theta}(\bx,\cdot) \) is trained to be idempotent given the 'empty' argument \( \bzer \). Of course, there is no guarantee of that when \( \bx \) does not come from the training distribution.
\subsection{Test-Time Training}
\label{sec:ttt}

In ZigZag~\cite{Durasov24a}, deviations from idempotence, as measured by the distance between the two predictions of Eq.~\ref{eq:zigzag}, are used to evaluate the accuracy of a prediction. In IT\(^3\) , we propose to go further and to minimize these deviations at {\it inference} time to compensate for potential domain shifts between training and test data. 

A naive way would be to minimize the loss function
\begin{align}
\mathcal{L}_{\text{TTT}} = \| f_{\theta}(\bx, f_{\theta}(\bx, \bzer)) - f_{\theta}(\bx, \bzer) \| \; , \label{eq:ttt-loss}
\end{align}
for all samples $\bx$ received as inference time. However, this can produce undesirable side effects.
For instance, if \( \by_0 = f_{\theta}(\bx, \bzer) \) is an incorrect prediction, minimizing \( \| \by_0 - \by_1 \| \) may cause \( \by_1 =  f_{\theta}(\bx, \by_0) \) to be pulled toward the incorrect \( \by_0 \), thereby magnifying the error. 
Another potential failure mode is to encourage $f_{\theta}(\bx,\cdot)$ to become the identity function, with is trivially idempotent. 

To prevent this, we modify the test-time training procedure as shown on the right side of Fig,~\ref{fig:method}.
We keep a copy of the model as it was at the end of the training phase, denoted as \( F = f_{\Theta} \), where \( \Theta \) are the weights obtained after the initial training of Sec.~\ref{ssec:zigzag-train}, which will not be updated further.
We then take the test-time loss to be
\begin{align}
\mathcal{L}_{\text{IT}^3} = \| F(\bx, f_{\theta}(\bx, \bzer)) - f_{\theta}(\bx, \bzer) \| \; ,
\end{align}
where \( f_{\theta} \) is the model being updated at test-time. Here, the first prediction \( \by_0 = f_{\theta}(\bx, \bzer) \) is computed as before, but the second one, \( \by_1 = F(\bx, f_{\theta}(\bx, \bzer)) \), is made using the frozen model \( F \).
By updating only \( f_{\theta} \) and keeping \( F \) fixed,
we ensure that \( \by_0 \) is adjusted to minimize the discrepancy with \( \by_1 \),
without pulling \( \by_1 \) toward an incorrect \( \by_0 \). A similar idea was employed in the IGN approach~\citep{Shocher24}
when meaningful predictions are required. After each TTT optimization iteration, the dynamic model $f_{\theta}$ is initialized with $\Theta$, ready for the next input.

Essentially, \itt{} extends the projection principle of Idempotent Generative Networks (IGN)~\cite{Shocher24}. IGNs 
map corrupted inputs onto the distribution of valid data by enforcing idempotence. Similarly, \itt{} projects OOD $(x, y)$ pairs onto the distribution of valid ones 
by iteratively refining the network’s internal representations. While only $y$ explicitly changes, 
every layer’s activations—functions of both $x$ and $y$—adjust to better fit the distribution of 
in-distribution representations, much like IGN corrects corrupted data by pulling it toward the natural image manifold. See detailed discussion in Appendix \ref{app:ign}.

% Essentially, $f$ is optimized to output $y_1$ s.t. when $F(x,\cdot)$ is applied to 
% it, it will remain the same. This would imply that the internal representation of $(x,y_1)$ is in the distribution 
% the model $F$ was trained on.

\subsection{Online IT\(^3\)}
\label{sec:online}

We introduce a variant of \itt{} for a \textit{different scenario}: Given data streams, where the distribution shifts continuously over time, in a continual learning setup, we modify \itt{} to operate in an online mode by not resetting $f_{\theta}$ back to $F$ after each TTT episode, as we did in Section~\ref{sec:ttt}.
We essentially assume that the distribution mostly shifts smoothly and, thus,  there is a good reason to believe that the current state of $f_{\theta}$ is a better 
initialization for the next TTT episode than the original $F$. This makes the model evolve over time.
In this scheme, it can happen that the performance of the model on data
from its original training decreases significantly, a phenomenon known as catastrophic
forgetting~\citep{Kirkpatrick17}. This is acceptable as the goal is to perform well on data at the present moment, rather than on past examples.

We make another modification in the second sequential application of the model, that is, when computing $F(\bx, f_{\theta}(\bx, \bzer))$.
Since the data keeps shifting, there is no reason to retain the frozen $F$ as an anchor indefinitely.
Over time, $f_{\theta}$ may diverge far from $F$, making it irrelevant. Relying on the old state of the model
would prevent the model from evolving efficiently. Replacing it with the current state of $f_{\theta}$ is out of 
the question, as it could causes collapse as described in Section~\ref{sec:ttt}. Instead, 
we need an anchor that is influenced by a reasonable amount
of data, yet evolves over time. Our solution is to replace \( F \) with an Exponential Moving Average (EMA) of the model \( f_{\theta} \), denoted as \( f_{\text{EMA}} \). This means \( f_{\text{EMA}} \) is a smoothed version of \( f_{\theta} \) over time.
The test-time loss in the online setting then becomes
\begin{align}
\mathcal{L}_{\text{online}} = \| f_{\text{EMA}}(\bx, f_{\theta}(\bx, \bzer)) - f_{\theta}(\bx, \bzer) \| \; .
\end{align}
By updating both \( f_{\theta} \) and \( f_{\text{EMA}} \) incrementally,
with \( f_{\text{EMA}} \) serving as a stable reference that changes more slowly,
the model adapts to gradual shifts without overfitting to noise or temporary anomalies.

\section{Experiments}
\label{sec:experiments}

We evaluate our approach across a diverse set of data types and tasks, including age prediction, image classification, and road segmentation in the visual domain, as well as aerodynamics prediction using 3D data and tabular data experiments.  In all these scenarios, we first train the model using the supervised approach of Section~\ref{ssec:zigzag-train} and then perform the test-time training of Section~\ref{sec:ttt}. For each task, we design an OOD test set for evaluation, that is, data drawn from a shifted distribution with respect to that of the training set. The OOD data is divided into several levels, with higher levels representing data that is progressively further from the training distribution. We evaluate our method for each level, presenting the results as bar plots for different batch sizes. After running the algorithm on a particular batch, we reset the model to its original, non-updated weights before evaluating the next batch.

% In each experiment, we observe how quickly the model's performance degrades as the level of OOD-ness increases. 

In our experiments, we compare our method against a non-optimized model to demonstrate the effectiveness of TTT approaches relative to a vanilla model, as well as other popular baselines. For the image classification task, we include the original TTT method and a newer more versatile approach, ActMAD~\cite{Mirza23}, which we described in Section~\ref{sec:related} and apply across all other setups. To further assess the effectiveness of our approach, we also evaluate all baselines, except the vanilla model, using different batch sizes. Across all scenarios, our method degrades more slowly than the baselines as the domain shift between training and testing data increases. Additionally, in Appendix~\ref{app:inference_time}, we provide inference time comparisons for the considered approaches.

% Unfortunately,  we have not found test-time any adaptation baselines matching the TTT problem setup for tasks other than image classification. Thus, we provide the comparative numbers in this case. For all others,  we compare against the performance of the vanilla non adaptive model. We systematically used published, common, strong models that are state-of-the-art or close to it. Across all scenarios, our method degrades slower than the baselines \pf{as the domain shift between training and testing data increases}.

\subsection{Tabular Data}

\begin{figure}[ht!]
    \centering
    % \begin{minipage}{.49\textwidth}
    %     \centering
    %     \includegraphics[width=\textwidth]{images/uci_ittt_results_mean.pdf}
    %     \vspace{-7mm}
    % \end{minipage}%
    % \hfill%
    \begin{minipage}{.49\textwidth}
        \centering
        \includegraphics[width=\textwidth]{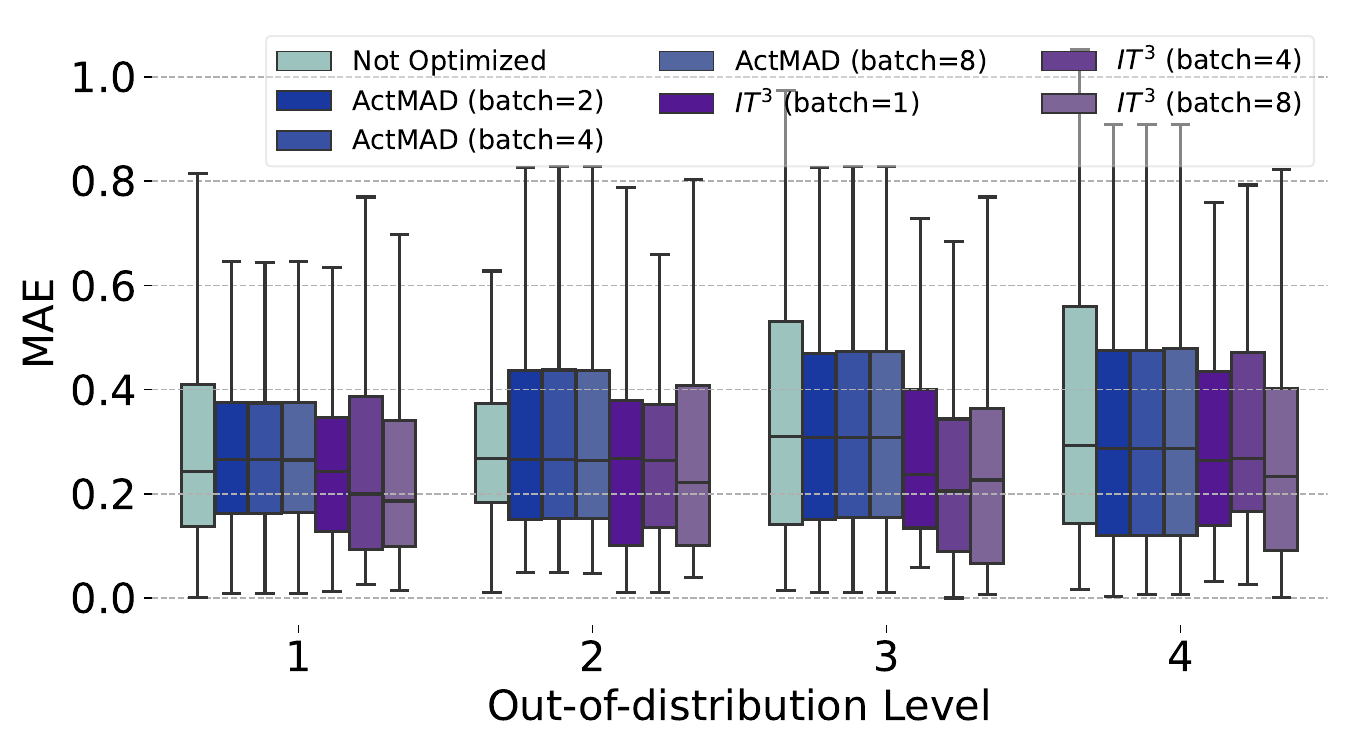}
        \vspace{-8mm}
    \end{minipage}
    \caption{\small \textbf{UCI Results on OOD inputs}: The plots illustrate the performance of \itt{} compared to other baselines across different OOD levels. The box plot for tabular data shows the distribution of MAE at various OOD levels, where $IT^3$ with different batch sizes ([\textcolor[rgb]{0.325,0.016,0.659}{\textbf{batch=1}}, \textcolor[rgb]{0.412,0.204,0.620}{\textbf{batch=4}}, \textcolor[rgb]{0.494,0.361,0.631}{\textbf{batch=8}}]) degrades less compared to the \textcolor[HTML]{96cac1}{\textbf{Not optimized}} baseline and \textcolor[rgb]{0.008,0.176,0.722}{\textbf{ActMAD}}. Larger batch sizes preserve performance more effectively.}
     \vspace{-3mm}
    % \caption{\small \textbf{UCI Results on OOD inputs}: The plots illustrate the performance of \itt{} compared to a vanilla model across different OOD levels. \textbf{Left}: The mean absolute error (MAE) shows that ITTT outperforms the vanilla model, retaining performance better as the data shifts further from the training distribution. \textbf{Right}: The box plot for car data shows the distribution of MAE at various OOD levels, where ITTT with different batch sizes ([\textcolor[HTML]{FFDAB9}{\textbf{batch=1}}, \textcolor[HTML]{C1BED6}{\textbf{batch=4}}, \textcolor[HTML]{EA8E83}{\textbf{batch=8}}]) degrades less compared to the \textcolor[HTML]{96cac1}{\textbf{Not optimized}} baseline. Larger batch sizes preserve performance more effectively.}
    \label{fig:uci_results}
\end{figure}

Tabular data consists of numerical features and corresponding continuous target values for regression tasks from the UCI tabular datasets~\citep{Bay20a}. They are widely used in machine learning research to benchmark regression models. In our case, we use The Boston Housing dataset describes housing prices in the suburbs of Boston, Massachusetts. It includes various features related to socioeconomic and geographical factors that influence housing prices. We take a test set and gradually apply random feature zeroing with increasing probabilities of 5\%, 10\%, 15\%, and 20\% (4 mentioned levels of OOD). This random feature dropping simulates out-of-distribution (OOD) data by progressively altering the input features, making the data less similar to the original training distribution. As the probability of feature dropping increases, the data becomes more OOD, which lowers the model's accuracy. The trained model is a simple Multi-Layer Perceptron (MLP) optimized using the Adam optimizer, and we observe that \itt{} consistently degrades less compared to other baselines across all OOD levels as depicted in Fig.~\ref{fig:uci_results}.

% \PF{It reads as though you are estimating uncertainty, which is not what we are trying to do here.}

% !TEX root = ../top.tex
% !TEX spellcheck = en-US

% \subsection{MNIST}

% \nd{We train the networks on MNIST~\citep{LeCun98a} and compute the accuracy and calibration metrics. We then use the uncertainty measure they produce to classify images from the test sets of MNIST and FashionMNIST~\citep{Xiao17a}  as being within the MNIST distribution or not to compute the OOD metrics introduced above. We use a standard architecture with four convolution and pooling layers, followed by fully connected layers with ReLU activations.}

\subsection{CIFAR}
\label{sec:cifar}

% !TEX root = ../top.tex
% !TEX spellcheck = en-US

\begin{figure*}[htbp]
  \centering
  \includegraphics[width=0.9\linewidth]{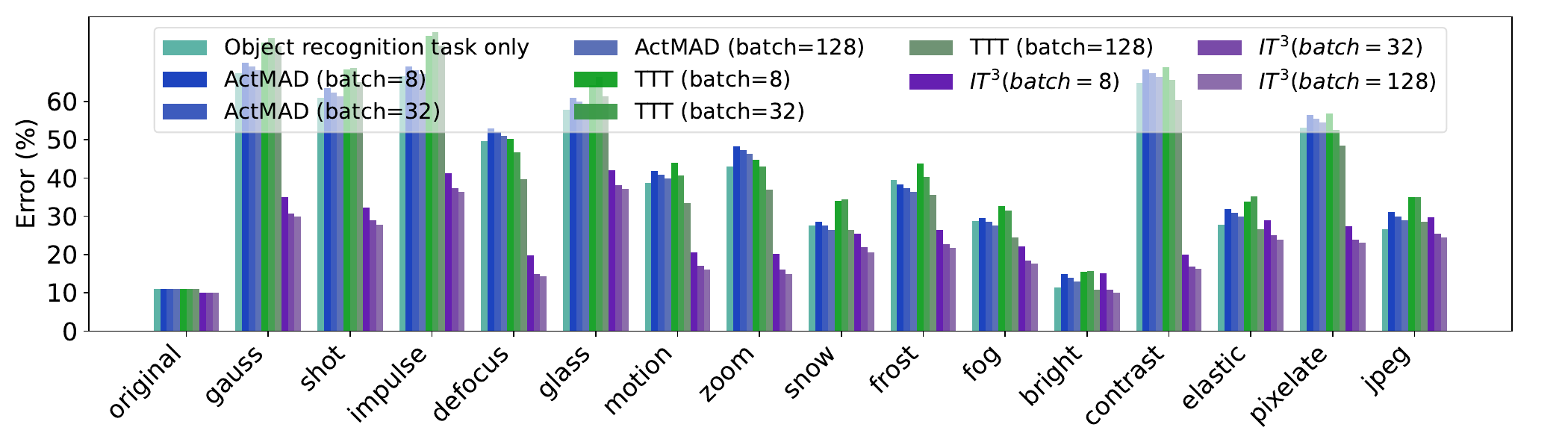}
  \vspace{-6mm}
   \caption{\small {\bf Test error (\%) on CIFAR-10-C with level 5 corruptions.} We compare our approaches, $IT^{3}$, with object recognition without self-supervision, TTT, and ActMAD. $IT^3$ improves over other baselines and higher batch size improves even further.}
  \label{fig:cifar_bars}
\end{figure*}

We conducted similar experiments using the CIFAR-10~\citep{Krizhevsky14} dataset, selecting CIFAR-C~\citep{Hendrycks18} as the out-of-distribution (OOD) data. CIFAR-C contains the same images as CIFAR-10 but with various common corruptions, such as Gaussian noise, blur, and contrast variations, simulating real-world conditions. These corruptions are applied at different severity levels, allowing us to evaluate how the model's performance degrades as the data shifts further from the original CIFAR-10 distribution. For this experiment, we used the \textit{Deep Layer Aggregation} (DLA)~\citep{Yu18f} network, known for its strong performance in image classification and robustness to overfitting. We trained the model according to the guidelines from the original DLA paper to ensure optimal results. Fig.\ref{fig:uci_results} shows the evaluation error on CIFAR-C at severity level 5 for different types of corruptions, following~\citep{Sun20c}. As shown, \itt{} outperforms other baselines, with higher batch sizes yielding the best results.
% In our basic setup, batch size of 1 does not work well. We did not explore the possibility of following \cite{Sun20c} creating a batch of augmented copies, as this would be a domain specific element, hurting the purity of our general method.
% For context, we add TTT results, while fully acknowledging that this is not a fair comparison as they have access to a single instance, making the batch size effectively 1, although augmented to a batch of size 32.

\subsection{Age Prediction}

% !TEX root = ../top.tex
% !TEX spellcheck = en-US

\newcommand{\imgwidth}{0.10\textwidth}
\begin{figure}[ht!]
    \centering
    \begin{minipage}{.48\textwidth}
        \centering
        \begin{tabular}{@{}c@{}c@{}c@{}c@{}c@{}c@{}c@{}c@{}c@{}c@{}}  % @{} removes extra padding between columns
            \includegraphics[width=\imgwidth]{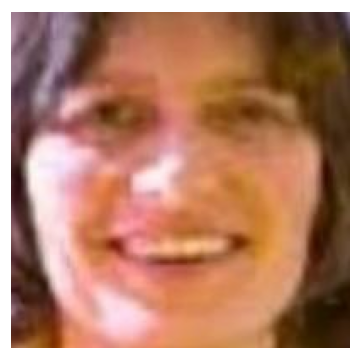} & 
            \includegraphics[width=\imgwidth]{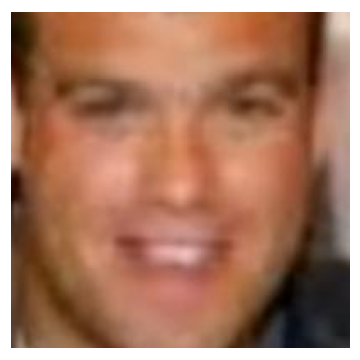} & 
            \includegraphics[width=\imgwidth]{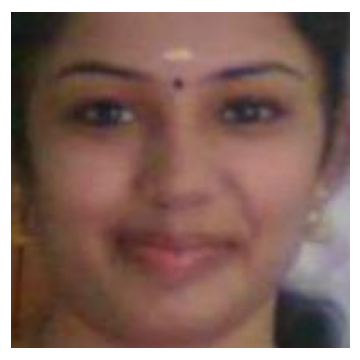} &
            \includegraphics[width=\imgwidth]{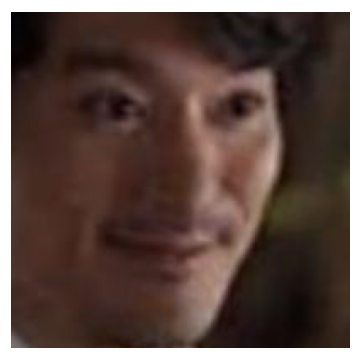} & 
            \includegraphics[width=\imgwidth]{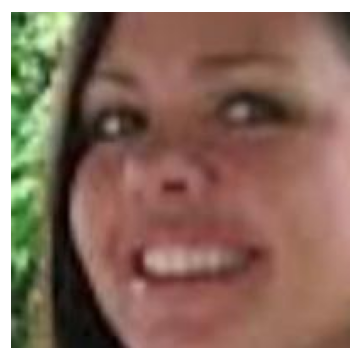} & 
            \includegraphics[width=\imgwidth]{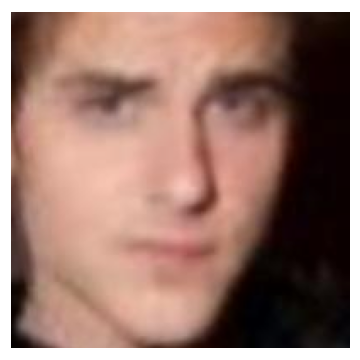} & 
            \includegraphics[width=\imgwidth]{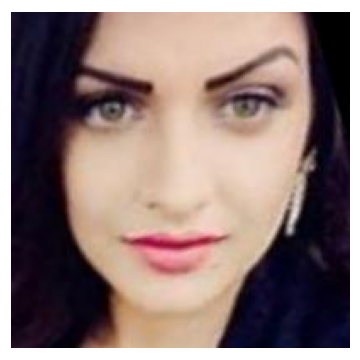} &
            \includegraphics[width=\imgwidth]{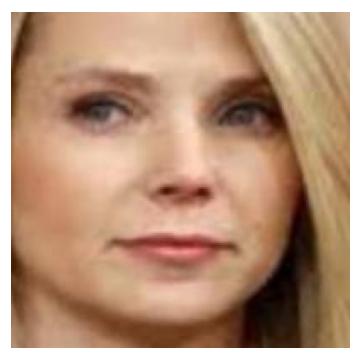} &
            \includegraphics[width=\imgwidth]{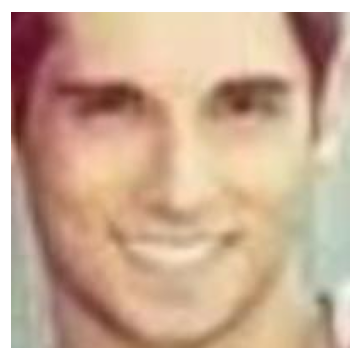} &
            \includegraphics[width=\imgwidth]{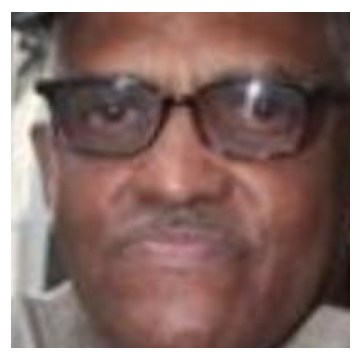}\\
           \includegraphics[width=\imgwidth]{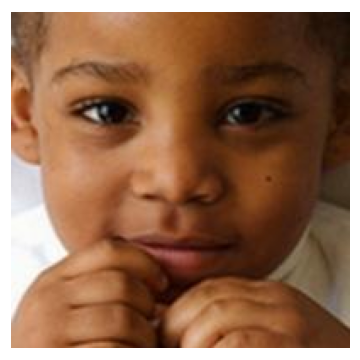} & 
            \includegraphics[width=\imgwidth]{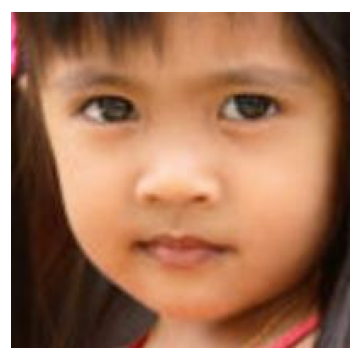} & 
            \includegraphics[width=\imgwidth]{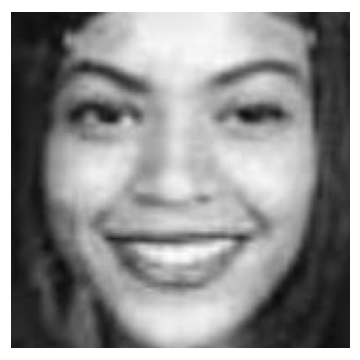} &
            \includegraphics[width=\imgwidth]{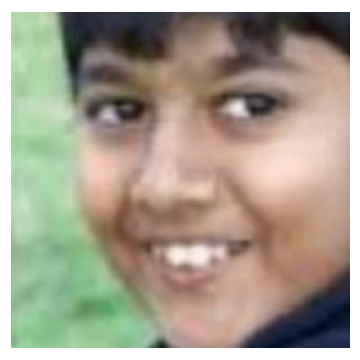} &
            \includegraphics[width=\imgwidth]{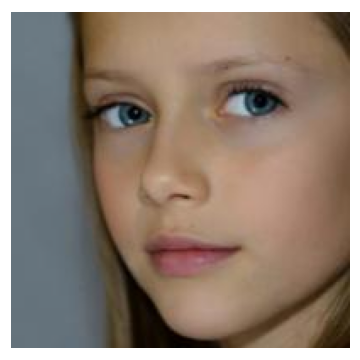} & 
            \includegraphics[width=\imgwidth]{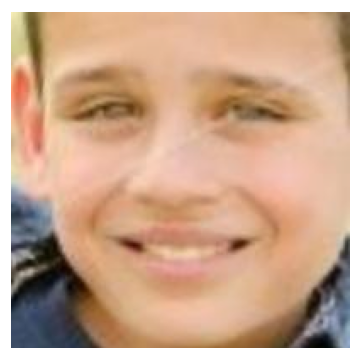} & 
            \includegraphics[width=\imgwidth]{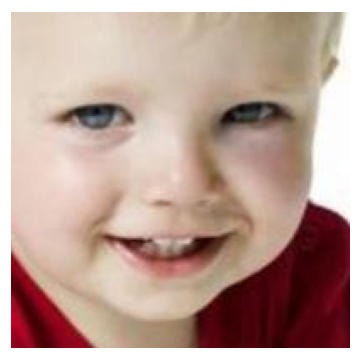} &
            \includegraphics[width=\imgwidth]{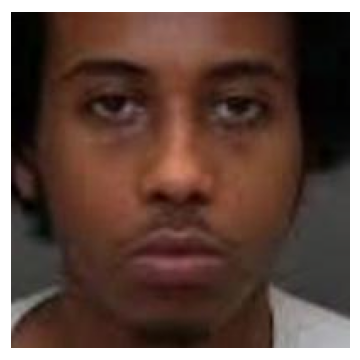} &
            \includegraphics[width=\imgwidth]{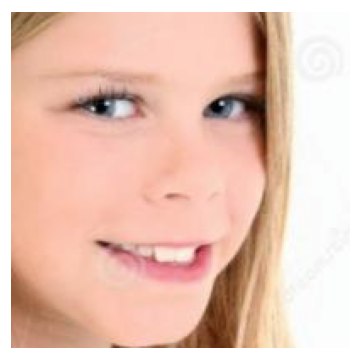} &
            \includegraphics[width=\imgwidth]{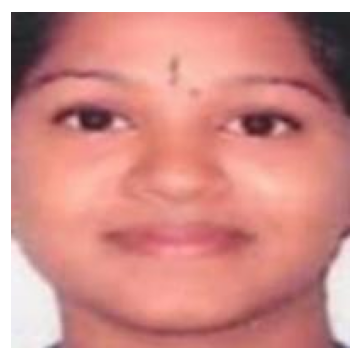} \\
            \includegraphics[width=\imgwidth]{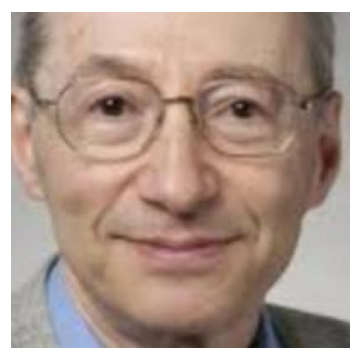} & 
            \includegraphics[width=\imgwidth]{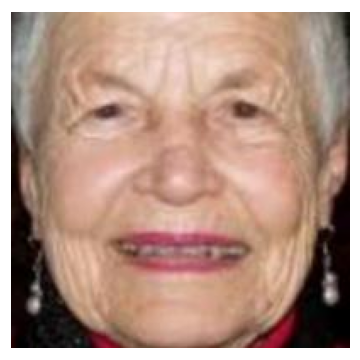} & 
            \includegraphics[width=\imgwidth]{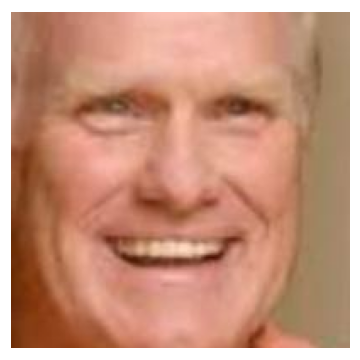} &
            \includegraphics[width=\imgwidth]{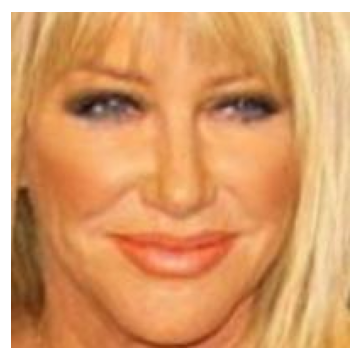} &
            \includegraphics[width=\imgwidth]{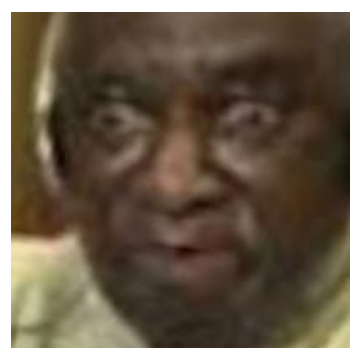} & 
            \includegraphics[width=\imgwidth]{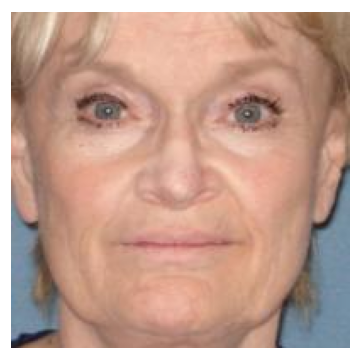} & 
            \includegraphics[width=\imgwidth]{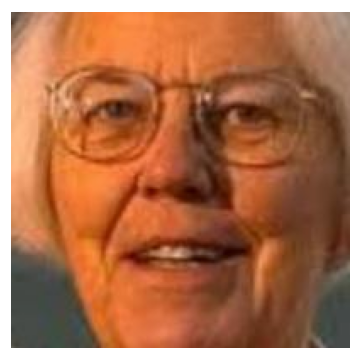} &
            \includegraphics[width=\imgwidth]{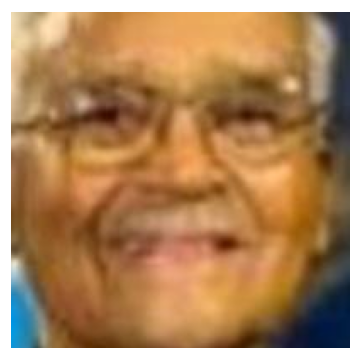} &
            \includegraphics[width=\imgwidth]{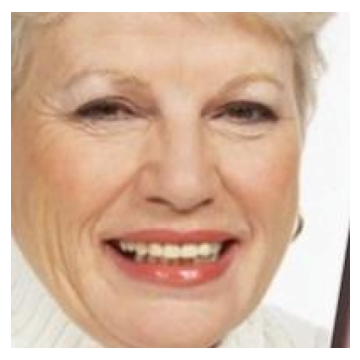} &
            \includegraphics[width=\imgwidth]{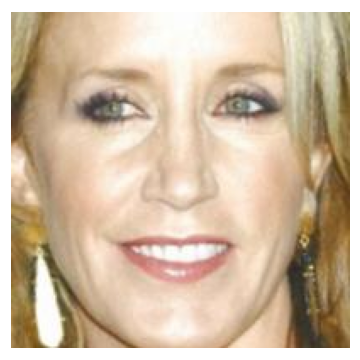} \\
        \end{tabular}
        \vspace{-4mm}
        \caption{\small {\bf Face Samples.} The \textbf{(top)} row shows training images of middle-aged individuals, while \textbf{(middle)} and \textbf{(bottom)} display images of older and younger individuals (OOD).}

        \label{fig:age_viz}

        % \vspace{1mm}

        % \includegraphics[width=\textwidth]{images/age_ittt_results_mean.pdf}
        % \vspace{-8mm}
        % \caption{\small {\bf Age  mean results on OOD shapes.} The plot compares \itt{} to a baseline, showing better performance retention as data shifts from the training distribution.}
        % \label{fig:age_mean}
        
    \end{minipage}%
    \hfill%
    \begin{minipage}{.48\textwidth}
        \centering
        \includegraphics[width=\textwidth]{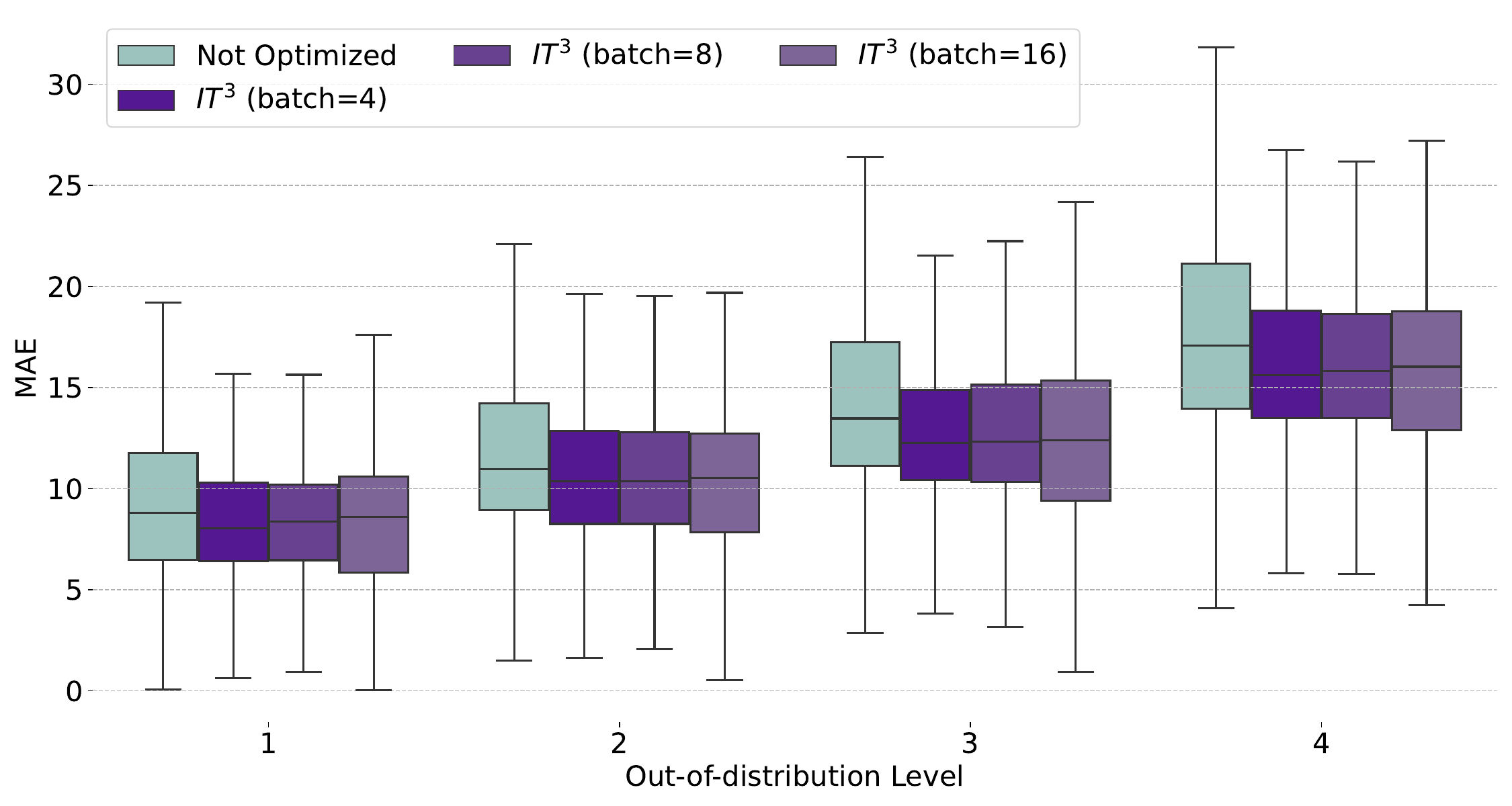}
        \vspace{-7mm}
        \caption{\small {\bf Age  boxplot results on OOD shapes.} \textcolor[HTML]{96cac1}{\textbf{Not optimized}} corresponds to a single model without TTT applied. \itt{} with [\textcolor[rgb]{0.325,0.016,0.659}{\textbf{batch=4}}, \textcolor[rgb]{0.412,0.204,0.620}{\textbf{batch=8}}, and \textcolor[rgb]{0.494,0.361,0.631}{\textbf{batch=16}}] represents our method at different batch sizes. As the data shifts further from the training distribution, our method degrades less, with larger batches preserving performance more effectively.}
         \vspace{-6mm}
        \label{fig:age_box}
    \end{minipage}
\end{figure}

% \textcolor[rgb]{0.325,0.016,0.659}{\textbf{batch=1}}, \textcolor[rgb]{0.412,0.204,0.620}{\textbf{batch=4}}, \textcolor[rgb]{0.494,0.361,0.631}{\textbf{batch=8}}

To experiment with image-based age prediction from face images, we use the UTKFace dataset~\citep{Zhifei17}, a large-scale collection containing tens of thousands of face images annotated with age information. The model is trained on face images of individuals aged between 20 and 60, while individuals younger or older than this range are considered out-of-distribution (OOD) (Fig.\ref{fig:age_viz}). The further the age is from the 20-60 interval, the higher the OOD level we assign to it. We use a ResNet-152 backbone with five additional linear layers and ReLU activations. This architecture delivers strong accuracy, outperforming the popular ordinal regression model CORAL~\citep{Cao20b} and matching other state-of-the-art methods~\citep{Berg21}. We train our model on the UTKFace training set (limited to individuals aged 20-60) and then run inference on faces at different OOD levels. Again, \itt{} significantly outperforms the non-optimized model, as shown in Fig.~\ref{fig:age_box}.

\subsection{Road Segmentation}

% !TEX root = ../top.tex
% !TEX spellcheck = en-US

\begin{figure}[ht!]
    \centering
    \begin{minipage}{.43\textwidth}
        \centering
        \includegraphics[width=\textwidth]{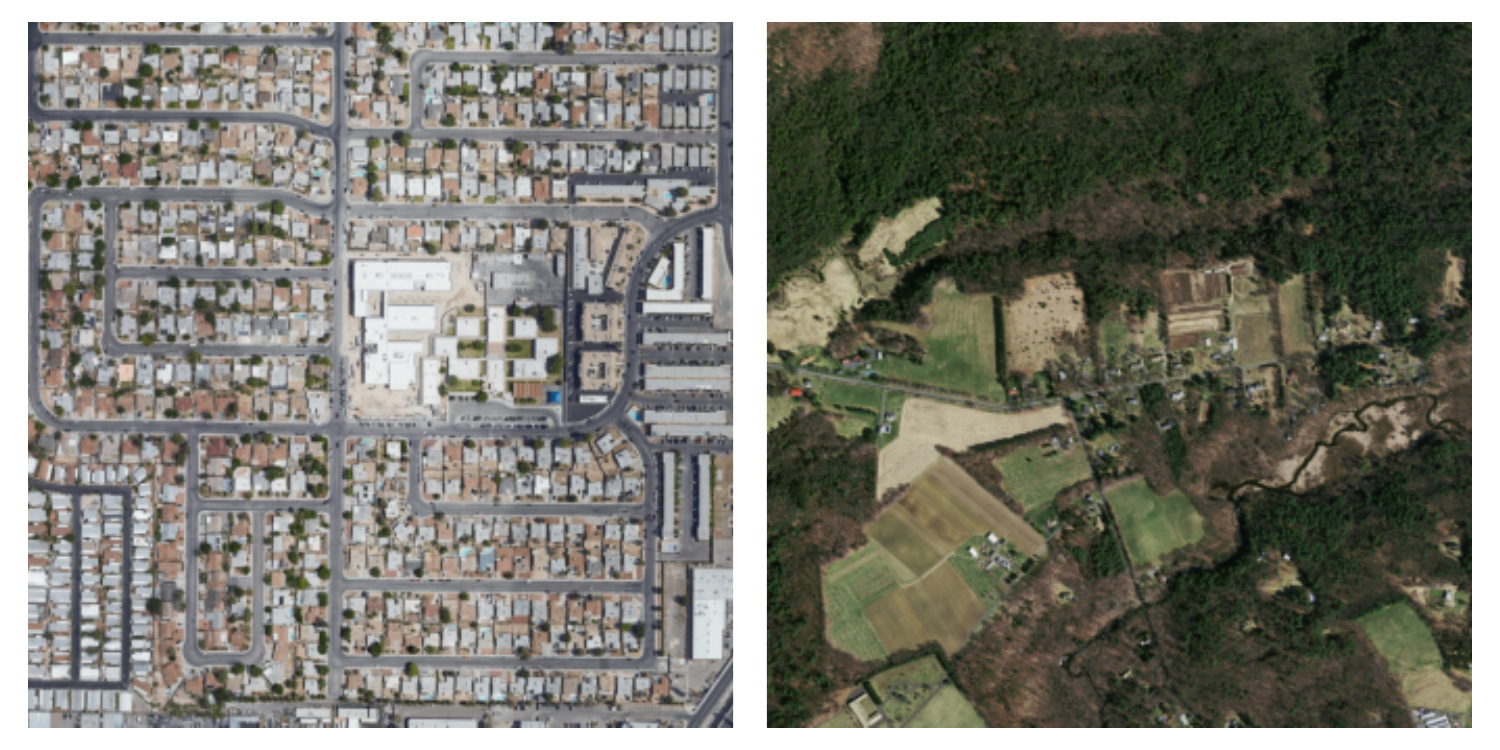}
        \vspace{-7mm}
        \caption{\small {\bf Road Samples.} 
        The roadTracer dataset \textbf{(left)} covers urban areas of six different countries while the Massachusetts dataset \textbf{(right)} primarily features rural neighborhoods along with some urban areas.
        }
        \label{fig:roads_ims}
    \end{minipage}%
    \hfill%
    \begin{minipage}{.48\textwidth}
        \centering
        \includegraphics[width=\textwidth]{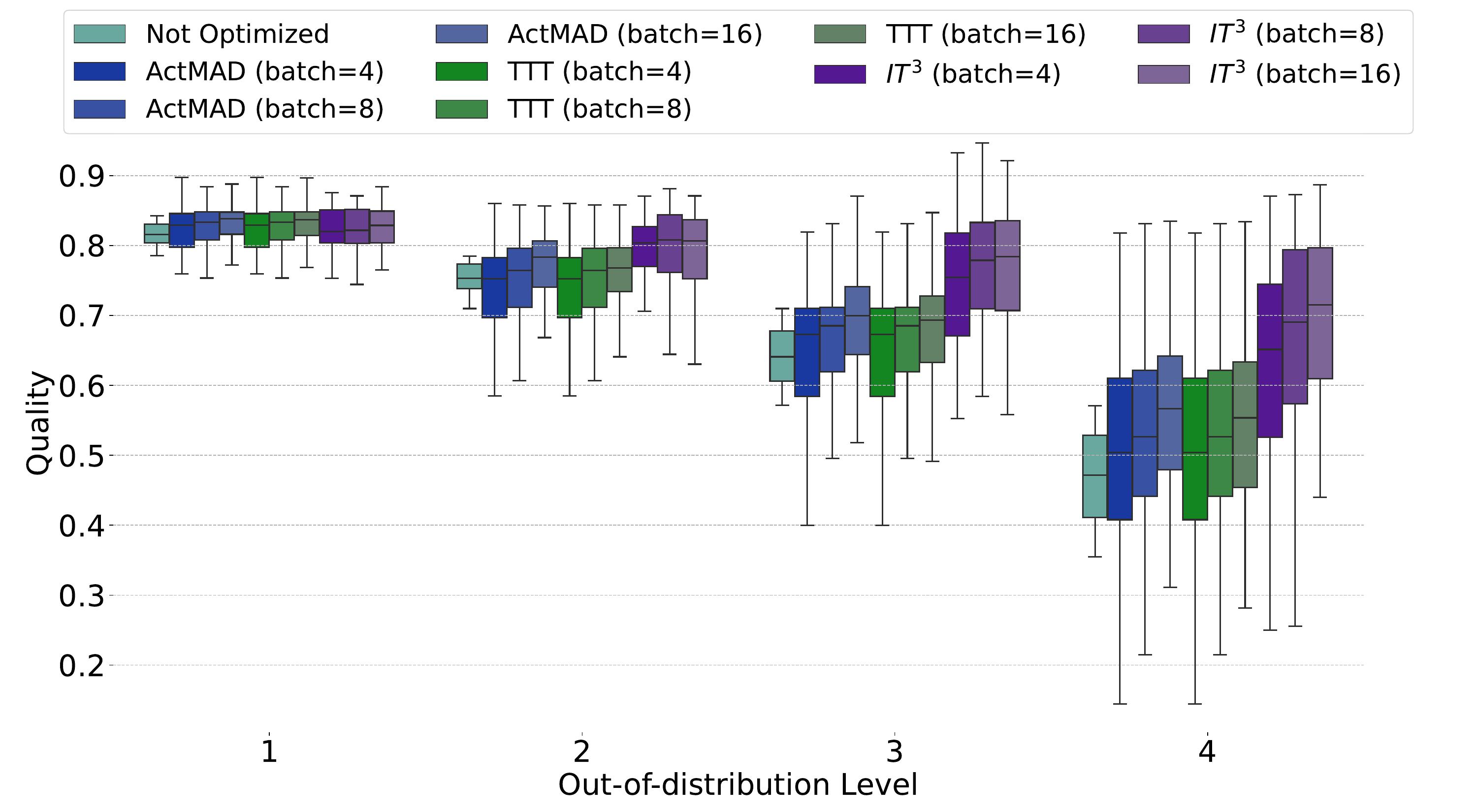}
         \vspace{-6mm}
        \caption{\small {\bf Roads results on OOD images.} \textcolor[HTML]{96cac1}{\textbf{Not optimized}} corresponds to a single model without TTT applied. \itt{} with [\textcolor[rgb]{0.325,0.016,0.659}{\textbf{batch=4}}, \textcolor[rgb]{0.412,0.204,0.620}{\textbf{batch=8}}, and \textcolor[rgb]{0.494,0.361,0.631}{\textbf{batch=16}}] represents our method at different batch sizes. As the data shifts further from the training distribution, our method degrades less compared to the \textcolor[HTML]{96cac1}{\textbf{Not optimized}}, \textcolor[rgb]{0.000,0.600,0.078}{\textbf{TTT}}, and \textcolor[rgb]{0.008,0.176,0.722}{\textbf{ActMAD}}, with larger batches preserving performance more effectively.}
         \vspace{-4mm}
        \label{fig:roads_box_results}
    \end{minipage}
\end{figure}

% \textcolor[rgb]{0.325,0.016,0.659}{\textbf{batch=1}}, \textcolor[rgb]{0.412,0.204,0.620}{\textbf{batch=4}}, \textcolor[rgb]{0.494,0.361,0.631}{\textbf{batch=8}}
% !TEX root = ../top.tex
% !TEX spellcheck = en-US

\begin{figure*}[ht!]
 % \vspace{-2mm}
  \centering
    \includegraphics[width=0.9\textwidth]{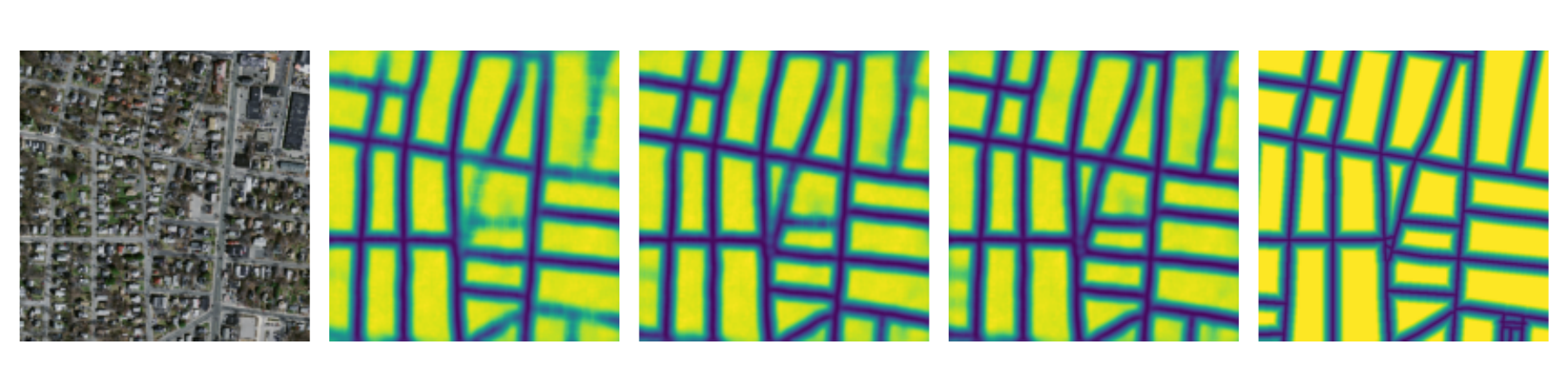} 
    \vspace{-8mm}
    \caption{\small {\bf Qualitative effect of \itt{} on Road Segmentation.} From left to right: (1) Original aerial image, (2) Output before optimization, (3) \itt{} output at the 5th iteration, (4) \itt{} output at the 15th iteration, and (5) Ground truth label. The segmentation quality improves significantly with \itt{} iterations, as observed in the progressively refined outputs at the 5th and 15th iterations.} 
     %\vspace{-5mm}
    \label{fig:road_iters}
\end{figure*}

Our method can be generalized to segmentation tasks as well. To demonstrate this, we consider the problem of road segmentation in aerial imagery using the RoadTracer dataset~\citep{Bastani18}. We train a DRU-Net~\citep{Wang19c}, on the RoadTracer dataset.

We perform OOD experiments using Massachusetts Road dataset~\citep{Mnih13} that primarily comprises rural neighborhoods, as depicted in Fig.~\ref{fig:roads_ims}. We sample 450 images, each with dimensions of $1500\times1500$ pixels and divide them into four groups based on the Mean Squared Error (MSE) of the segmentation outputs, effectively creating different levels of distributional shift within the sampled set. We then further train the network on these OOD subsets using the ZigZag method \citep{Durasov24a}.

We evaluate road segmentation performance by using \textit{Correctness}, \textit{Completeness} and \textit{Quality} (CCQ) metric~\citep{Wiedemann98} which is a popular metric to evaluate delineation performance. The \textit{Correctness}, \textit{Completeness} and \textit{Quality} are equivalent to precision, recall and intersection-over-union, where the definition of a true positive has been relaxed from spatial coincidence of prediction and annotation to co-occurrence within a distance of 5 pixels. As shown in Fig.\ref{fig:roads_box_results}, \itt{} significantly improves performance on OOD images (see Fig.\ref{fig:ttt_segmentation_comparison} for more qualitative examples).

\subsection{Aerodynamics Prediction}
\label{ssec:aero}

% !TEX root = ../top.tex
% !TEX spellcheck = en-US

\begin{figure}[ht!]
    \centering
    \begin{minipage}{.48\textwidth}
        \centering
        \includegraphics[width=\textwidth]{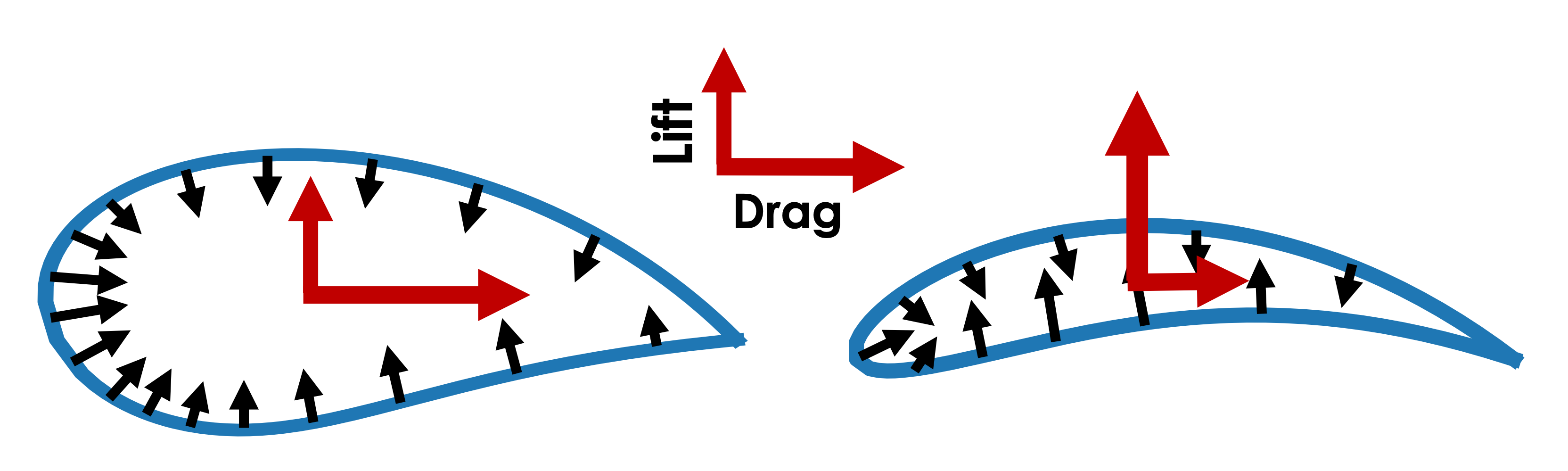}
        \vspace{-6mm}
        \caption{\small {\bf Airfoil Samples.} 
         Training and testing profiles (\textbf{left}) show reasonable aerodynamics, while OOD samples (\textbf{right}) feature rare, high lift-to-drag shapes. Black arrows indicate pressure, and red lines show lift and drag.
        }
        \label{fig:airfoil_viz}

        % \vspace{2mm}

        % \includegraphics[width=\textwidth]{images/airfoils_ittt_results_mean.pdf}
        % \vspace{-7mm}
        % \caption{\small {\bf Cars mean error on OOD shapes.} The plot compares ITTT to the vanilla model, showing better performance retention as data shifts from the training distribution.}
        % \label{fig:airfoil_mean}
        
    \end{minipage}%
    \hfill%
    \begin{minipage}{.48\textwidth}
        \centering
        \includegraphics[width=\textwidth]{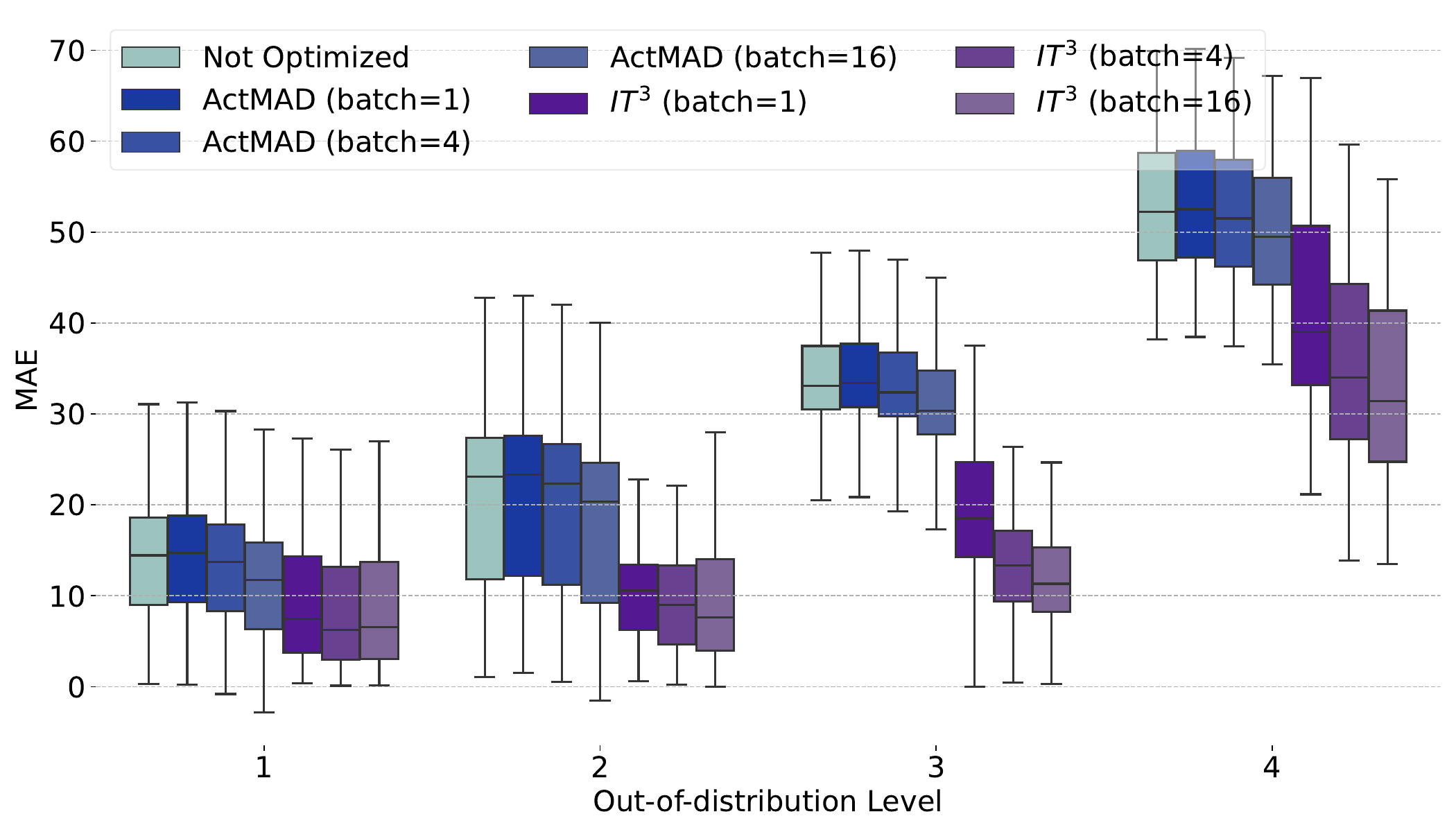}
        \vspace{-8mm}
        \caption{\small {\bf Airfoil results on OOD shapes.} \textcolor[HTML]{96cac1}{\textbf{Not optimized}} corresponds to a single model without TTT applied. \textbf{ITTT} with [\textcolor[rgb]{0.325,0.016,0.659}{\textbf{batch=1}}, \textcolor[rgb]{0.412,0.204,0.620}{\textbf{batch=4}}, and \textcolor[rgb]{0.494,0.361,0.631}{\textbf{batch=16}}] represents our method at different batch sizes. As the data shifts further from the training distribution, our method degrades less compared to the \textcolor[HTML]{96cac1}{\textbf{Not optimized}} and \textcolor[rgb]{0.008,0.176,0.722}{\textbf{ActMAD}}, with larger batches preserving performance more effectively.}
          \vspace{-4mm}
        \label{fig:airfoil_box}
    \end{minipage}
\end{figure}
% !TEX root = ../top.tex
% !TEX spellcheck = en-US

\begin{figure}[ht!]
    \centering
    \begin{minipage}{.48\textwidth}
        \centering
        \includegraphics[width=\textwidth]{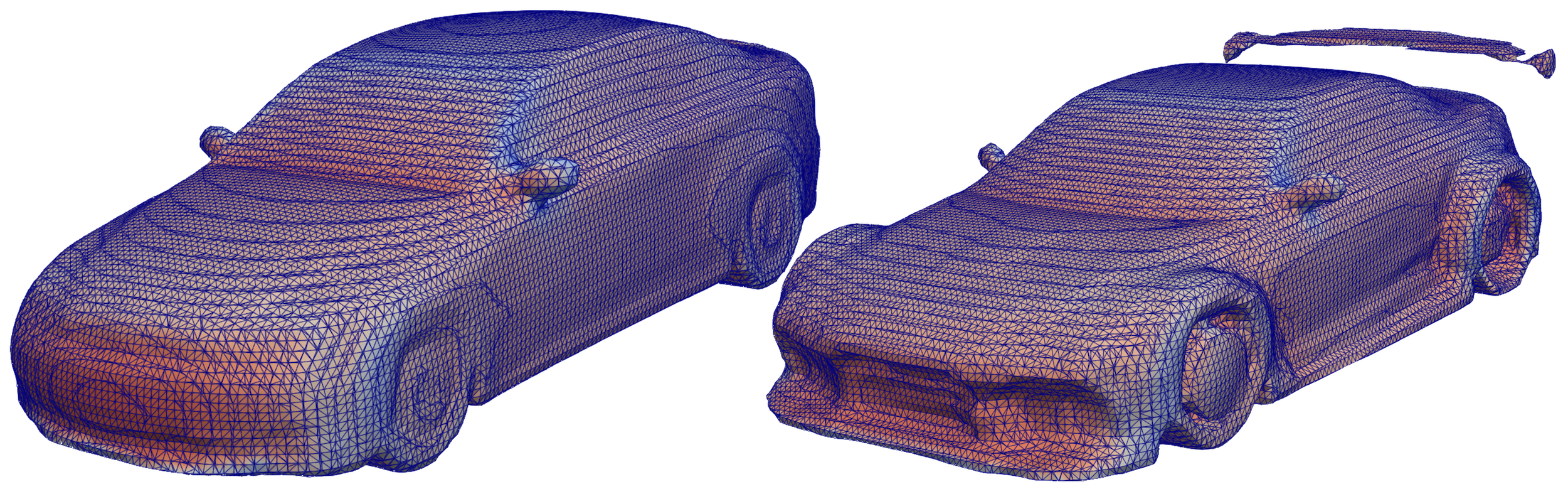}
          \vspace{-4mm}
        \caption{\small {\bf Car Samples.} 
        The car dataset comprises many regular vehicles \textbf{(left)} and a few streamlined ones \textbf{(right)}, which we treat as being out-of-distribution. Red and blue denote high and low pressures respectively.
        }
        \label{fig:car-viz}

        % \vspace{2mm}

        % \includegraphics[width=\textwidth]{images/cars_ittt_results_mean.pdf}
        % \vspace{-7mm}
        % \caption{\small {\bf Cars mean error on OOD shapes.} The plot compares ITTT to the vanilla model, showing better performance retention as data shifts from the training distribution.}
        % \label{fig:car-mean}
        
    \end{minipage}%
    \hfill%
    \begin{minipage}{.48\textwidth}
        \centering
        \includegraphics[width=\textwidth]{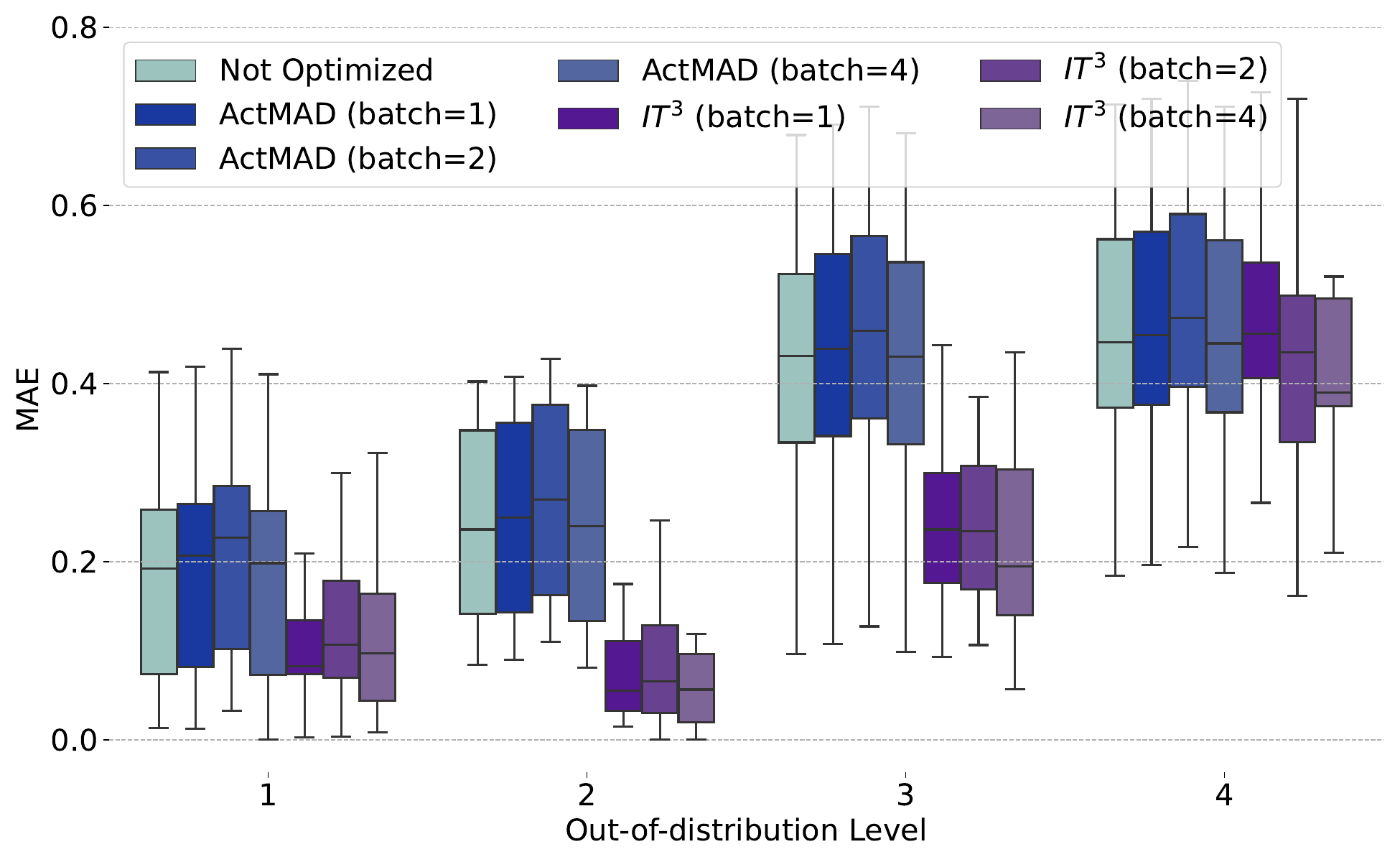}
        \vspace{-8mm}
        \caption{\small {\bf Car results on OOD shapes.} \textcolor[HTML]{96cac1}{\textbf{Not optimized}} corresponds to a single model without TTT applied. \textbf{ITTT} with [\textcolor[rgb]{0.325,0.016,0.659}{\textbf{batch=1}}, \textcolor[rgb]{0.412,0.204,0.620}{\textbf{batch=2}}, and \textcolor[rgb]{0.494,0.361,0.631}{\textbf{batch=4}}] represents our method at different batch sizes. As the data shifts further from the training distribution, our method degrades less compared to the \textcolor[HTML]{96cac1}{\textbf{Not optimized}} baseline and \textcolor[rgb]{0.008,0.176,0.722}{\textbf{ActMAD}}, with larger batches preserving performance more effectively.}
         \vspace{-6mm}
        \label{fig:car-box}
    \end{minipage}
\end{figure}

\paragraph{Wings.}
Our method is versatile and can handle various types of data. To illustrate this, we generated a dataset of 2,000 wing profiles, as depicted in Fig.\ref{fig:airfoil_viz}, by sampling the widely used NACA parameters~\citep{Jacobs37}. We used the XFoil simulator~\citep{Drela89} to compute the pressure distribution along each profile and estimate its lift-to-drag coefficient, a crucial indicator of aerodynamic performance. The resulting dataset consists of wing profiles $\bx_i$, represented by a set of 2D nodes, and the corresponding scalar lift-to-drag coefficient $\by_i$ for $1 \leq i \leq 2000$.

% !TEX root = ../top.tex
% !TEX spellcheck = en-US

% \begin{table}[t]
% \caption{\textbf{Qualitative result for Road Segmentation:} We report evaluation metrics for the road segmentation task performed on Massachusetts dataset, which acts as an OOD dataset since the initial network was trained on the RoadTracer dataset. The results suggest \itt{} enhances the segmentation performance compared to the initial network. Additionally, online \itt{} siginificantly outperforms offline \itt{}.}
% \label{tab:roads_ccq}
% \vskip 0.15in
% \begin{center}
% \begin{small}
% \begin{sc}
% \begin{tabular}{lccc}
% \toprule
% Method            &   \textit{Corr}   &   \textit{Comp}   &   \textit{Quality}    \\
% \midrule
% Not Optimized     &   $55.7$           &   $44.3$           &   $39.5$               \\

% \itt{} (batch=4)  &   $55.7$           &   $49.1$           &   $46.4$               \\

% \itt{} (batch=8)  &   $58.1$           &   $52.0$           &   $48.5$               \\

% \itt{} (batch=16) &   $57.3$           &   $52.7$           &   $48.7$               \\

% \itt{} (online)   &   $\textbf{77.5}$ &   $\textbf{79.8}$ &  $\textbf{69.8}$      \\
% \bottomrule
% \end{tabular}
% \end{sc}
% \end{small}
% \end{center}
% \vskip -0.1in
% \end{table}

\begin{table*}[t]
% \caption{\textbf{Qualitative result for Road Segmentation:} We report evaluation metrics for the road segmentation task performed on Massachusetts dataset, which acts as an OOD dataset since the initial network was trained on the RoadTracer dataset. The results suggest \itt{} enhances the segmentation performance compared to the initial network. Additionally, online \itt{} significantly outperforms offline \itt{}.}
\caption{\textbf{Qualitative result for Online \itt{}.} Evaluation metrics for the road segmentation task \textbf{(left)}, airfoils lift-to-drag prediction \textbf{(middle)}, and car drag prediction \textbf{(right)}.  Online \itt{} enhances performance compared to the original model and significantly outperforms offline \itt{}.}
\vspace{-3mm}
\label{tab:roads_ccq}
\vskip 0.15in
\begin{center}
\begin{small}
\begin{sc}
\begin{minipage}{0.435\linewidth}
\centering
\begin{tabular}{lccc}
\toprule
Method            &   \textit{Corr}   &   \textit{Comp}   &   \textit{Quality}    \\
\midrule
Not Optimized     &   $55.7$           &   $44.3$           &   $39.5$               \\

\itt{} (batch=4)  &   $55.7$           &   $49.1$           &   $46.4$               \\

\itt{} (batch=8)  &   $58.1$           &   $52.0$           &   $48.5$               \\

\itt{} (batch=16) &   $57.3$           &   $52.7$           &   $48.7$               \\

\itt{} (online)   &   $\textbf{77.5}$ &   $\textbf{79.8}$ &  $\textbf{69.8}$      \\
\bottomrule
\end{tabular}
\end{minipage}%
\hfill
\begin{minipage}{0.25\linewidth}
\centering
\begin{tabular}{lc}
\toprule
Method            &   \textit{MAE}     \\
\midrule
Not Optimized     &   $38.2$      \\

\itt{} (batch=1)  &   $37.6$  \\

\itt{} (batch=4)  &   $37.5$   \\

\itt{} (batch=16) &   $37.4$    \\

\itt{} (online)   &   $\textbf{34.1}$ \\
\bottomrule
\end{tabular}
\end{minipage}
\hfill
\begin{minipage}{0.25\linewidth}
\centering
\begin{tabular}{lc}
\toprule
Method            &   \textit{MAE}     \\
\midrule
Not Optimized     &   $0.501$      \\

\itt{} (batch=1)  &   $0.446$  \\

\itt{} (batch=2)  &   $0.424$   \\

\itt{} (batch=4) &   $0.412$    \\

\itt{} (online)   &   $\textbf{0.385}$ \\
\bottomrule
\end{tabular}
\end{minipage}
\end{sc}
\end{small}
\end{center}
\vspace{-6mm}
\label{tab:roads_ccq}
\end{table*}

We selected the top 5\% of shapes, based on their lift-to-drag ratio, as out-of-distribution (OOD) samples. The OOD levels were determined using the ground truth lift-to-drag ratio, where higher OOD levels correspond to more aerodynamically streamlined shapes. The training set includes shapes with lift-to-drag values ranging from 0 to 60, with anything beyond this threshold considered OOD and excluded from training. We then trained a Graph Neural Network (GNN) composed of 25 GMM~\citep{Monti17} layers, featuring ELU activations~\citep{Clevert15} and skip connections~\citep{He16a}, to predict the lift-to-drag coefficient $\by_i$ from the profile $\bx_i$, following the approach of~\citep{Remelli20b, Durasov24a}. As with previous experiments, \itt{} significantly improves performance on OOD shapes and provides more accurate predictions compared to other baselines, as shown in Fig.~\ref{fig:airfoil_box}.

\parag{Cars.}
As for wings, we experimented with 3D car models from a subset of the ShapeNet dataset~\citep{Chang15}, which contains car meshes suitable for CFD simulations, as depicted in Fig.~\ref{fig:car-viz}. The experimental protocol was the same as for the wing profiles, except we used OpenFOAM~\citep{Jasak07} to estimate drag coefficients and employed a more sophisticated network to predict them from the triangulated 3D car meshes.

To predict drag associated to a triangulated 3D car, we utilize similar model to airfoil experiments but with increased capacity. Instead of twenty five GMM layers, we use thirty five and also apply skip-connections with ELU activations.
Final model is being trained for 100 epochs with Adam optimizer and $10^{-3}$ learning rate. As with airfoils experiments, \itt{} significantly improves performance on OOD shapes and provides more accurate predictions compared to other baselines, as shown in Fig.~\ref{fig:car-box}.

% \newpage
\subsection{Online \itt{}}
We test our proposed online variation on several tasks. \textit{Assuming a data stream online scenario rather than the previous setup}. Naturally, when the distribution remains constant (although shifted from the training distribution) we expect superior results w.r.t. the offline setup, as our model keeps being trained on the new distribution. 
A more effective way to test constant adaptation over time is to use a continuously changing distribution. We test \itt{} on an increasing corruption / OOD level. In all cases, the online \itt{} performs significantly better than the basic anchored variation.

\textbf{Road segmentation.} Building upon our previous road segmentation experiments, we further evaluate the effectiveness of online \itt{}. In the online \itt{} setup, OOD samples are ranked based on their mean squared error (MSE) loss when passed through the vanilla network. We begin by selecting the samples with low MSE loss, as these are closer to the training distribution given the network's strong performance on them. Gradually, we introduce samples with progressively higher MSE loss, smoothly shifting between distributions and thereby allowing the model to adapt effectively to a range of OOD samples. As in previous experiment, we use DRU-Net trained on the RoadTracer dataset as vanilla model and 890 images are sampled from Massachusetts dataset as OOD images.

Firstly, the vanilla network is tested on the Massachusetts dataset without any additional fine-tuning. We then apply online \itt{} during inference to adapt the model to the OOD distribution as new data is presented. We evaluate the segmentation performance using the \textit{Correctness}, \textit{Completeness}, and \textit{Quality} metrics, as described previously. Table~\ref{tab:roads_ccq} (left) summarizes the results. The application of \itt{} improved the performance over the initial network and the online \itt{} method significantly outperforms the offline \itt{}.

\textbf{Aerodynamics.} We conducted online experiments for airfoils and cars lift-to-drag prediction.
We set the data stream such that OOD shapes appear with increasing aerodynamic properties, modeling a continuous domain shift in the data. As with the segmentation results, the online version significantly outperforms both the offline version and the original network, as shown in Tabs.\ref{tab:roads_ccq} (middle and right).

\subsection{ImageNet}

\begin{figure*}[t!]
    \centering
    \begin{minipage}{.99\textwidth}
        \centering
        \includegraphics[width=\textwidth]{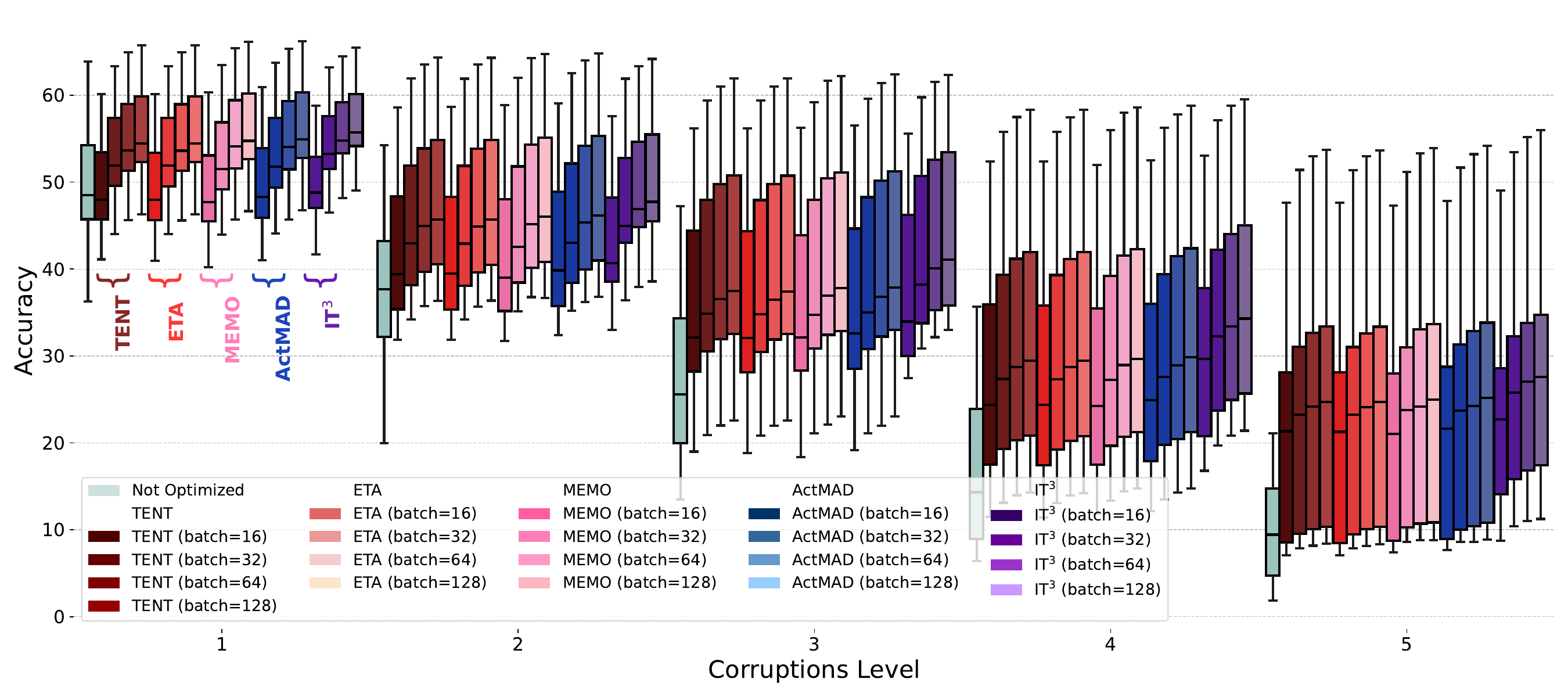}
        \vspace{-7mm}
    \end{minipage}
    \caption{\small \textbf{Test accuracy (\%) on ImageNet-C across 5 corruption levels.} This plot compares our method, $IT^3$, with popular adaptation approaches including TENT, ETA, MEMO, and ActMAD on the ImageNet-C dataset. $IT^3$ consistently outperforms all baselines across all corruption severity levels and batch sizes, with performance further improving at higher batch sizes.}
    \vspace{-3mm}
    \label{fig:imagenet_results}
\end{figure*}

    % ([\textcolor[rgb]{0.325,0.016,0.659}{\textbf{batch=1}}, \textcolor[rgb]{0.412,0.204,0.620}{\textbf{batch=4}}, \textcolor[rgb]{0.494,0.361,0.631}{\textbf{batch=8}}]) some description goes some description goes some description goes some description goes some description goes some description goes some description goes some description goes some description goes some description goes

\begin{figure}[ht!]
    \centering
    \begin{minipage}{.49\textwidth}
        \centering
        \includegraphics[width=\textwidth]{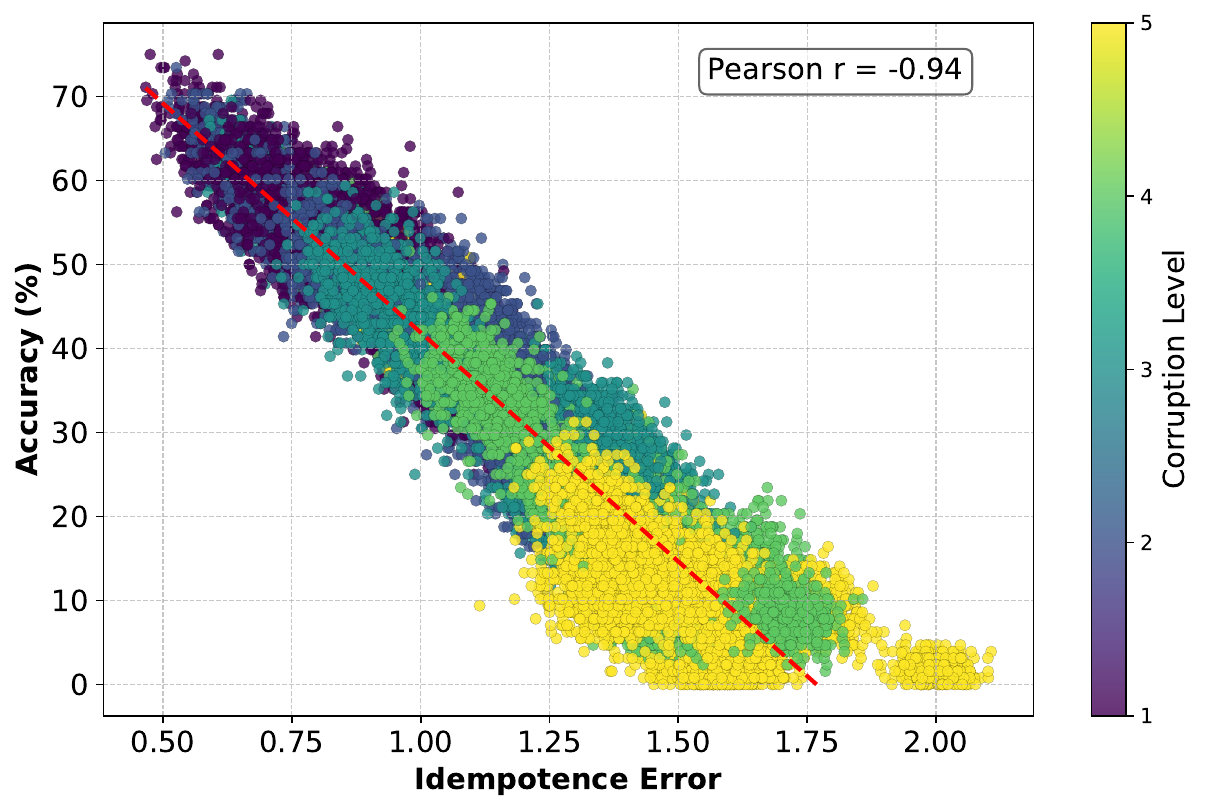}
        \vspace{-8mm}
    \end{minipage}
    \caption{\small \textbf{Accuracy vs Idempotence on Corrupted ImageNet}: This plot demonstrates a strong correlation between model idempotence and performance. Each point corresponds to one inference batch, with accuracy and idempotence error computed per batch. The color indicates the corruption level from ImageNet-C for the respective batch. The Pearson correlation between idempotence error and accuracy is $-0.94$, indicating a strong negative trend.
    }
     \vspace{-3mm}
    \label{fig:acc_vs_idempotence}
\end{figure}

ImageNet-C~\cite{Hendrycks18} consists of ImageNet~\cite{Krizhevsky12} test images corrupted using the same transformations as CIFAR-10/100C (Sec.~\ref{sec:cifar}). We used it for large-scale classification experiments, evaluating performance across 15 corruption types and different severity levels. For our setup, we employed a standard ResNet-18~\cite{He16a} and followed the Pytorch training protocol~\cite{Paszke17}. In addition to our previous experiments, we included several widely used baselines on this benchmark: TENT~\cite{wang2020tent}, ETA~\cite{niu2022efficient}, and MEMO~\cite{zhang2022memo}.

The results on ImageNet-C (each bar represents the average accuracy across 15 corruption types) are shown in Table~\ref{fig:imagenet_results}. Our method outperforms all other approaches across all corruption levels and batch sizes, and significantly surpasses the original baseline model. As previously observed, larger inference batch sizes improve performance for all methods. These results demonstrate that our approach is also effective in large-scale data scenarios.

\paragraph{Accuracy vs Idempotence.} Similar to the results in Fig.~\ref{fig:idempotence_effect}, Fig.~\ref{fig:acc_vs_idempotence} illustrates the correlation between model performance and idempotence error. This supports the core idea of our method: optimizing idempotence during inference can improve performance. As shown, idempotence error exhibits a strong negative correlation with accuracy. This observation further reinforces the conclusions of~\cite{Shocher24} and~\cite{Durasov24a}, which highlight that idempotence error across multiple predictions is a strong indicator of model performance—a key motivation behind our approach.

% !TEX root = ../top.tex
% !TEX spellcheck = en-US

%\section{Limitations}
%
%While global, \itt{} lacks domain expertise. While we were able to demonstrate favorable result for \itt{} in all the domains we experimented with, we can't rule out the possibility to implement domain specific methods based on self-supervision that can outperform \itt{} for that domain. We found that for some domains it is hard to apply \itt{} on single instances without also using additions that require domain expertise or access to training data. This is most common in domains where the information within a single input is limited. Combining \itt{} with domain specific methods may remove these limitations, at the cost of generality. 
%
%
%\section{Conclusion}
%
%We have proposed an approach to test-time-training that relies on enforcing idempotence as new samples are being considered to effectively handle domain shifts. The method is generic and we have demonstrated that it is effective in a wide range of domains without requiring domain-specific knowledge, which sets it apart from other state-of-the-art methods.
%
%In future work we plan to pursue the challenge of realistic online continual learning, where there is no pre-training at all and the data arrives in streams, sometimes with labels and sometimes not. We believe \itt{} can be adapted to such a setup across many different streaming modalities, which would make it extremely useful in real-world scenarios.

\section{Conclusions, Limitations, and Future Work}

We have proposed an approach to test-time-training relying on enforcing idempotence as new samples are being considered. This effectively handles domain shifts and method is generic. We have demonstrated that it is effective in a wide range of domains without requiring domain-specific knowledge, which sets it apart from state-of-the-art methods. 

The flip side is that \itt{} lacks domain expertise. In some cases, it is hard to apply \itt{} to single instances without additional conditions. This is most common in domains where information within a single input is limited. Combining \itt{} with domain-specific methods may remove these limitations, which we will explore in future work.

\section*{Impact Statement}
This paper introduces \itt{}, a new approach that enables more adaptable machine-learning models by refining their predictions for out-of-distribution data at test time. Through this on-the-fly adaptation, \itt{} can reduce training costs, data requirements, and model size, making advanced AI methods more broadly accessible. The technique thus offers a step toward more efficient, flexible deployment of deep learning in real-world scenarios, where conditions often shift beyond the original training domain.
% This paper presents work whose goal is to advance the field of 
% Machine Learning. There are many potential societal consequences 
% of our work, none which we feel must be specifically highlighted here.

\bibliography{bib/string,bib/vision,bib/learning,bib/robotics,bib/stats,bib/cfd,bib/misc,bib/new}
\bibliographystyle{icml2025}

%%%%%%%%%%%%%%%%%%%%%%%%%%%%%%%%%%%%%%%%%%%%%%%%%%%%%%%%%%%%%%%%%%%%%%%%%%%%%%%
%%%%%%%%%%%%%%%%%%%%%%%%%%%%%%%%%%%%%%%%%%%%%%%%%%%%%%%%%%%%%%%%%%%%%%%%%%%%%%%
% APPENDIX
%%%%%%%%%%%%%%%%%%%%%%%%%%%%%%%%%%%%%%%%%%%%%%%%%%%%%%%%%%%%%%%%%%%%%%%%%%%%%%%
%%%%%%%%%%%%%%%%%%%%%%%%%%%%%%%%%%%%%%%%%%%%%%%%%%%%%%%%%%%%%%%%%%%%%%%%%%%%%%%
\newpage
\appendix
\onecolumn

\section{Inference Time Comparison}

% \subsection{Inference Time Comparison}
\label{app:inference_time}

Our method typically requires only 1--3 optimization steps, keeping the overall cost comparable to other well-known TTT methods. Below, we provide a comparison of inference times on out-of-distribution data for three approaches: the base model without optimization, the state-of-the-art TTT method ActMAD, and \itt{}. As shown, while our method introduces no significant overhead compared to ActMAD, it remains computationally efficient while achieving substantial improvements in performance. A similar observation can be made about memory consumption, as reported in Tab.~\ref{tab:mem_cons} for the case of batch size 128 on ImageNet-C, showing peak memory reserved (in GB) using \texttt{torch.cuda.max\_memory\_reserved()}.

\begin{table}[h]
    \centering
    \begin{minipage}[t]{0.48\textwidth}
        \centering
        \caption{Inference Time Comparison (OOD Airfoils)}
        \begin{tabular}{lccc}
            \toprule
            \textbf{Method} & \textbf{Base Model} & \textbf{ActMAD} & $\mathbf{IT^3}$ \\
            \midrule
            Inference Time (↓) & $1\times$ & $3\times$ & $4\times$ \\
            \bottomrule
        \end{tabular}
    \end{minipage}
    \hfill
    \begin{minipage}[t]{0.48\textwidth}
        \centering
        \caption{Inference Time Comparison (OOD Cars)}
        \begin{tabular}{lccc}
            \toprule
            \textbf{Method} & \textbf{Base Model} & \textbf{ActMAD} & $\mathbf{IT^3}$ \\
            \midrule
            Inference Time (↓) & $1\times$ & $4\times$ & $5\times$ \\
            \bottomrule
        \end{tabular}
    \end{minipage}
\end{table}

\begin{table}[h]
    \centering
    \begin{minipage}[t]{0.45\textwidth}
        \centering
        \caption{Inference Time Comparison (OOD Roads)}
        \begin{tabular}{lccc}
            \toprule
            \textbf{Method} & \textbf{BASE} & \textbf{ActMAD} & $\mathbf{IT^3}$ \\
            \midrule
            Inference Time (↓) & $1\times$ & $4.5\times$ & $6\times$ \\
            \bottomrule
        \end{tabular}
    \end{minipage}
    \hfill
    \begin{minipage}[t]{0.53\textwidth}
        \centering
        \caption{Memory Consumption (OOD ImagetNet), GPU Gb}
        \begin{tabular}{l@{\hskip 4pt} c@{\hskip 4pt} c@{\hskip 4pt} c@{\hskip 4pt} c@{\hskip 4pt} c@{\hskip 4pt} c@{\hskip 4pt}}
            \toprule
            \textbf{Method} & \textbf{BASE} & \textbf{TENT} & \textbf{MEMO} & \textbf{ETA} & \textbf{ActMAD} & $\mathbf{IT^3}$ \\
            \midrule
            Memory (↓) & $4.5$ & $4.8$ & $13.5$ & $4.9$ & $7.2$ & $7.4$ \\
            \bottomrule
        \label{tab:mem_cons}
        \end{tabular}
    \end{minipage}
    
\end{table}

\section{Extended Discussion: Relating \itt{} to IGN}
\label{app:ign}

In this section, we elaborate on how \itt{} generalizes the 
\emph{projection principle} of Idempotent Generative Networks (IGN) 
to a supervised test-time training setting. We show that both 
approaches use \emph{idempotence}---repeated applications of the 
network function should yield the same result---as a way to 
``project'' off-manifold inputs onto a learned manifold of valid data. 
While IGN enforces idempotence directly on all possible inputs, 
\itt{} enforces it primarily on training data but adapts 
on-the-fly at test time to handle out-of-distribution (OOD) samples.

\vspace{0.75em}
\noindent\textbf{The Projection Principle in IGN.}
\vspace{0.35em}

\noindent
IGN~\citep{Shocher24} learns 
\(\,g_{\theta}:\mathcal{Z}\to\mathcal{X}\), mapping from a 
\emph{source} distribution \(\mathcal{P}_{z}\) (e.g.\ Gaussian noise) 
to a \emph{target} distribution \(\mathcal{P}_{x}\subset\mathcal{X}\) 
(e.g.\ natural images). It imposes:
\[
  g_{\theta}\bigl(g_{\theta}(z)\bigr)\;=\;g_{\theta}(z)
  \quad\forall\,z\in\mathcal{Z},
\]
so a second application of \(g_{\theta}\) makes no change. This 
idempotence implies that once an off-manifold \(z\) is mapped to 
\(\,g_{\theta}(z)\), it must already lie on the manifold 
\(\{x:g_{\theta}(x)=x\}\). In effect, 
\[
  z\;\mapsto\;g_{\theta}(z)\;\in\;
  \Bigl\{x:\,g_{\theta}(x)=x\Bigr\}.
\]
One can interpret this as a \emph{projection}:
a drift or energy measure 
\(\;\delta_{\theta}(x)\!=\!\|\,g_{\theta}(x)-x\,\|\) 
vanishes (\(\delta_{\theta}(x)=0\)) if and only if \(x\) 
already lies on that manifold. Enforcing \(g_{\theta}(g_{\theta}(z)) 
= g_{\theta}(z)\) ensures \(\delta_{\theta}(g_{\theta}(z))=0\). 
Hence, after one forward pass, the corrupted or noisy input is 
``pulled'' onto the learned data manifold, 
and repeated applications do not alter it further.

\vspace{0.75em}
\noindent\textbf{Idempotence in \itt{}: Pairwise Function.}
\vspace{0.35em}

\noindent
\itt{} deals with a supervised model 
\[
  f_{\theta}: \mathcal{X}\times\mathcal{Y}
  \;\to\;\mathcal{Y},
\]
where \(\bx\in\mathcal{X}\) is an input and \(\by\in\mathcal{Y}\) 
its desired output. The training set 
\(\{(\bx_i,\by_i)\}\) spans an in-distribution 
\(\mathcal{P}_{x,y}\). During training, \itt{} enforces:
\begin{enumerate}
\item 
  \(f_{\theta}(\bx,\by)=\by\) for training pairs 
  \((\bx,\by)\). Thus, each real pair is a fixed point.
\item 
  \(f_{\theta}(\bx,\bzer)\approx\by\), using a ``neutral'' 
  label \(\bzer\) to predict \(\by\).
\end{enumerate}
Combining these yields:
\[
   f_{\theta}\Bigl(\bx,\;f_{\theta}(\bx,\bzer)\Bigr)
   \;=\;
   f_{\theta}(\bx,\bzer),
\]
an idempotence condition parallel to IGN’s 
\(g_{\theta}(g_{\theta}(z))=g_{\theta}(z)\). One may define 
a drift-like measure
\[
   \Delta_{\theta}(\bx) \;=\;
   \Bigl\|\,
      f_{\theta}\!\bigl(\bx,\,f_{\theta}(\bx,\bzer)\bigr)
      \;-\;
      f_{\theta}(\bx,\bzer)
   \Bigr\|.
\]
When \(\bx\) is in-distribution, training 
makes \(\Delta_{\theta}(\bx)=0\). If \(\bx\) is OOD, 
\(\Delta_{\theta}(\bx)>0\) initially. 
\textbf{Test-time adaptation} then updates \(\theta\) on-the-fly 
to push \(\Delta_{\theta}(\bx)\) closer to zero, thereby 
restoring idempotence.

\vspace{0.75em}
\noindent\textbf{Subtlety: Internal Representations Projection}
\vspace{0.35em}

\noindent
A natural question arises: 
\emph{If the OOD variable \(\bx\) itself stays fixed, how can 
\((\bx,\by)\) become ``on distribution''?} The answer is that, 
inside the network layers, \(\bx\) and \(\by\) jointly produce 
hidden representations. Although \(\bx\) does not physically 
change, the \emph{way} \(\bx\) participates in the representation 
\textit{does} change once \(\by\) and the model parameters \(\theta\) 
are updated. Thus, even if \(\bx\) is not from the training distribution, 
the pair \(\bigl(\bx,\widehat\by\bigr)\) can enter a region of 
representation space that \emph{matches} valid training pairs. 
Formally, each layer of \(f_{\theta}\) has activations that depend 
on both \(\bx\) and \(\by\). By adjusting \(\theta\) (but freezing 
the outer function call) so that 
\[
   f_{\theta}\!\Bigl(\bx,\;f_{\theta}(\bx,\bzer)\Bigr)
   = 
   f_{\theta}(\bx,\bzer),
\]
we effectively \emph{project} \(\bigl(\bx,\bzer\bigr)\) 
into the manifold 
\(\bigl\{(\bx,\by):f_{\theta}(\bx,\by)=\by\}\) 
\emph{within} the network’s internal representation. 
Hence, even though \(\bx\) remains the same, 
the final pair \((\bx,\widehat\by)\) is ``valid'' 
in the sense that repeated applications are stable.

\vspace{0.75em}
\noindent\textbf{Conclusion: \itt{} Also ``Projects'' Off-Manifold Pairs.}
\vspace{0.35em}

\noindent
In IGN, once trained, any input \(z\in\mathcal{Z}\) 
maps to \(g_{\theta}(z)\) on the real-image manifold 
(\(g_{\theta}(x)=x\)). In \itt{}, a new OOD pair 
\((\bx,\bzer)\) is adapted so that 
\(\bigl(\bx,f_{\theta}(\bx,\bzer)\bigr)\) belongs to the set 
\(\{(\bx,\by):f_{\theta}(\bx,\by)=\by\}\). From an internal 
representation viewpoint, this \emph{pulls} the OOD pair 
onto the manifold of valid \((\bx,\by)\) relations. 
Thus, \itt{} \emph{extends} IGN’s core idea of 
``learned idempotent projection'' to a supervised test-time 
training paradigm. Despite leaving \(\bx\) intact, 
the final output indeed corresponds to a consistent 
\((\bx,\by)\)-pair on the model’s manifold, 
much like IGN pulls corrupted noise into the real-data manifold.

\section*{Appendix C: Detailed Elaboration: Relation between Adaptation and Idempotence}

In this appendix, we provide a detailed explanation of the rationale behind the idempotence loss used in IT$^3$. Our aim is to demonstrate how the discrepancy between recursive model outputs quantifies the out-of-distribution (OOD) uncertainty and why actively minimizing this discrepancy drives the model toward idempotence with respect to its auxiliary input.

\subsection*{1. Training via the ZigZag Approach}
Following \cite{Durasov24a}, the network is modified to accept an auxiliary input. During training, for each training pair \((x,y)\), the model \( f \) is trained to satisfy:
\begin{equation}
f(x,0) \approx y \quad \text{and} \quad f(x,y) \approx y.
\end{equation}
This is achieved by minimizing a composite loss:
\begin{equation}
L_{\mathrm{train}} = \| f(x,0) - y \| + \| f(x,y) - y \|.
\end{equation}
This training strategy ensures that incorporating the auxiliary input (either the true label or a “don’t know” signal) does not hurt the primary task performance while establishing a consistency property in the network.

\subsection*{2. Test-Time Recursive Inference and Uncertainty Measurement}
At test time, the network is applied recursively:
\begin{align}
y_0 &= f(x,0), \label{eq:y0} \\
y_1 &= f(x,y_0). \label{eq:y1}
\end{align}
We define the uncertainty loss as:
\begin{equation}
L_{\mathrm{IT^3}}(x) = \| y_1 - y_0 \|.
\end{equation}
The rationale is as follows:
\begin{enumerate}
    \item If \(x\) is in-distribution, then \(y_0 \approx y\), and since the network is trained so that \(f(x,y) \approx y\), we have \(y_1 \approx y_0\). Therefore, \(L_{\mathrm{IT^3}}(x)\) is small.
    \item If \(x\) is OOD, then \(y_0\) is unlikely to approximate the true label. In this case, the pair \((x,y_0)\) is not a valid input as per training, leading \(y_1\) to be unpredictable and significantly different from \(y_0\), resulting in a large \(L_{\mathrm{IT^3}}(x)\).
\end{enumerate}
Thus, the magnitude of \( \| y_1 - y_0 \| \) serves as a proxy for prediction certainty.

\subsection*{3. From Uncertainty to Idempotence}
We now elaborate on how the uncertainty loss translates into enforcing idempotence of \( f(x,\cdot) \) for a given \( x \). Recall that a function is idempotent if
\begin{equation}
f(f(x)) = f(x).
\end{equation}
For our setting, consider the function \( g(y) = f(x,y) \). The desired idempotence condition becomes:
\begin{equation}
g(g(0)) = g(0),
\end{equation}
or equivalently,
\begin{equation}
f(x, f(x,0)) = f(x,0).
\end{equation}
Thus, we can rewrite the uncertainty loss as:
\begin{equation}
L_{\mathrm{IT^3}}(x) = \| f(x, f(x,0)) - f(x,0) \|.
\end{equation}
Minimizing \(L_{\mathrm{IT^3}}(x)\) drives the network toward the condition that repeated application of \( f(x,\cdot) \) does not change the output. When \( L_{\mathrm{IT^3}}(x)=0 \), the function \( f(x,\cdot) \) is idempotent given the input \( x \). This self-consistency indicates that the network’s output is aligned with the in-distribution manifold.

\subsection*{4. Addressing Optimization Challenges}
Minimizing the idempotence loss is not trivial. Naively reducing \(L_{\mathrm{IT^3}}\) can lead to pitfalls such as reinforcing an erroneous prediction \(y_0\). As discussed in \cite{Shocher24}, directly optimizing the idempotence loss induces two gradient pathways:
\begin{enumerate}
    \item A desirable pathway that updates \(f(x,0)\) toward the correct in-distribution manifold.
    \item An undesirable pathway that may cause the manifold to expand, thereby including an incorrect \(f(x,0)\).
\end{enumerate}
To counteract the latter, our method avoids passing gradients through the second application of \(f\) by using a frozen copy of the network (or updating it via an exponential moving average). Specifically, we compute:
\begin{equation}
y_1 = F(x, f(x,0)),
\end{equation}
and redefine the loss as:
\begin{equation}
L_{\mathrm{IT^3}} = \| F(x, f(x,0)) - f(x,0) \|.
\end{equation}
This decoupling ensures that only the first prediction \(y_0\) is adapted during test-time training, thus preventing error reinforcement and ensuring that the minimization of the loss indeed pulls \(f(x,\cdot)\) toward idempotence.

\begin{figure*}[!h]
\centering
\includegraphics[width=\textwidth]{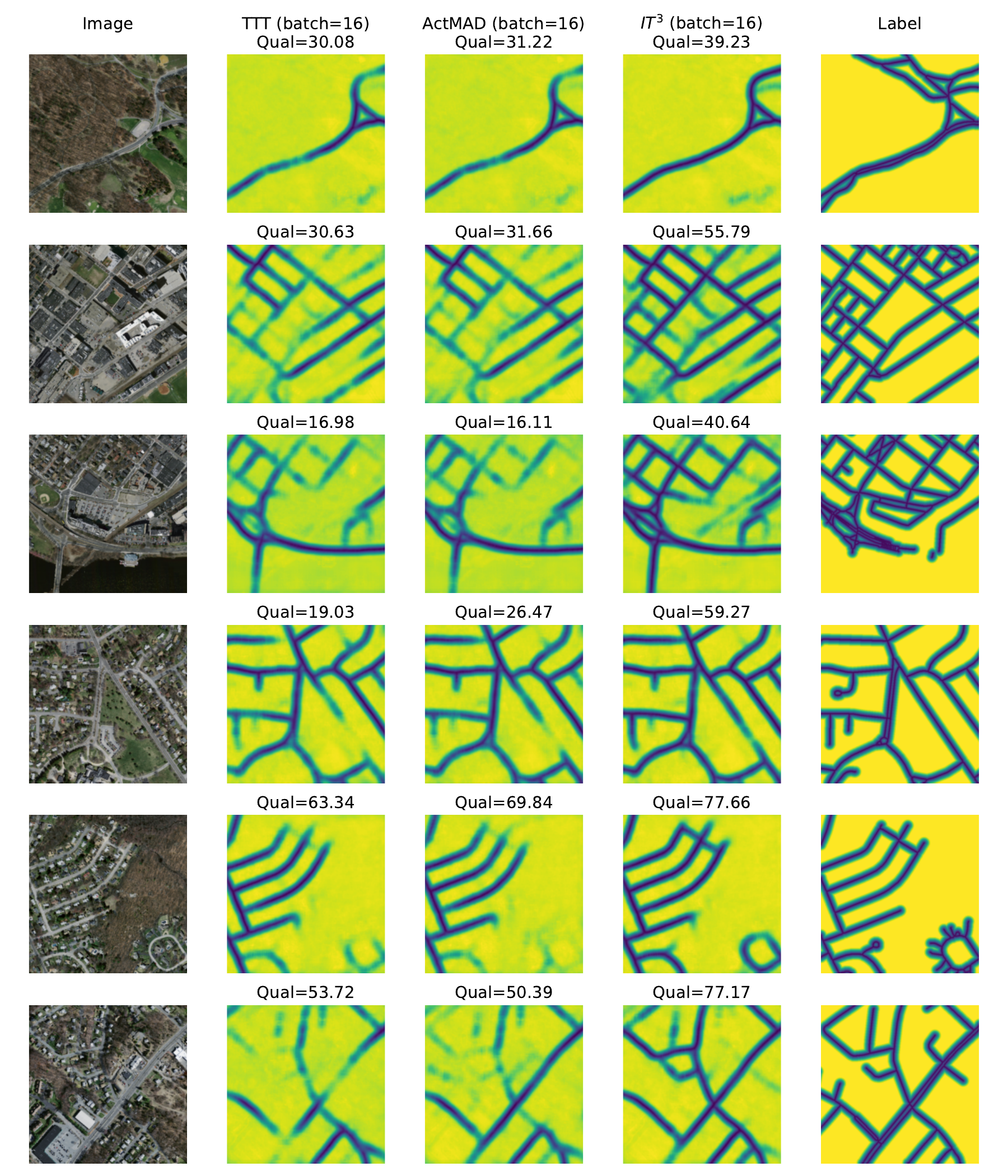}
\caption{ \textbf{Comparison of different Test-Time Training methods on segmentation tasks.} This plot shows the predictions of various TTT methods on an aerial segmentation task. As can be seen, our approach consistently enhances prediction quality and outperforms other methods.}
\label{fig:ttt_segmentation_comparison}
\end{figure*}

\subsection*{5. Summary}
By minimizing the loss
\begin{equation}
L_{\mathrm{IT^3}} = \| F(x, f(x,0)) - f(x,0) \|,
\end{equation}
we enforce the condition
\begin{equation}
f(x, f(x,0)) \approx f(x,0),
\end{equation}
i.e., \( f(x,\cdot) \) becomes idempotent given the input \( x \). This idempotence is a critical indicator that the input is aligned with the training distribution. When \( L_{\mathrm{IT^3}} \) is small, the prediction is self-consistent and reliable; when it is large, it signals that the input is likely OOD. Consequently, actively minimizing this loss through test-time training refines the network's prediction and enhances its robustness to distribution shifts.

\vspace{1em}
\noindent For further details on related projection perspectives, please refer to Appendix B.


% You can have as much text here as you want. The main body must be at most $8$ pages long.
% For the final version, one more page can be added.
% If you want, you can use an appendix like this one.  

% The $\mathtt{\backslash onecolumn}$ command above can be kept in place if you prefer a one-column appendix, or can be removed if you prefer a two-column appendix.  Apart from this possible change, the style (font size, spacing, margins, page numbering, etc.) should be kept the same as the main body.
%%%%%%%%%%%%%%%%%%%%%%%%%%%%%%%%%%%%%%%%%%%%%%%%%%%%%%%%%%%%%%%%%%%%%%%%%%%%%%%
%%%%%%%%%%%%%%%%%%%%%%%%%%%%%%%%%%%%%%%%%%%%%%%%%%%%%%%%%%%%%%%%%%%%%%%%%%%%%%%

\end{document}